\documentclass{article} 

\usepackage[a4paper, total={6in, 8in}]{geometry}

\usepackage{amsmath,amsfonts}
\usepackage[algo2e]{algorithm2e} 
\usepackage{array}
\usepackage{tikz-network}
\usepackage{caption}
\usepackage{subcaption}
\usepackage{booktabs}
\usepackage{textcomp}
\usepackage{stfloats}
\usepackage{url}
\usepackage{verbatim}
\usepackage{graphicx}
\usepackage[export]{adjustbox}
\usepackage{cite}
\usepackage{blindtext}
\usepackage{wrapfig}
\usepackage{rotating}

\usepackage{pdflscape}

\newcolumntype{P}[1]{>{\centering\arraybackslash}p{#1}}
\newcolumntype{M}[1]{>{\centering\arraybackslash}m{#1}}

\usepackage{pgfplots}
\pgfplotsset{compat=1.15}
\usepackage{mathrsfs}

\usepackage{caption}
\usepackage{enumitem}

\usepackage{tikz}
\usetikzlibrary{arrows.meta,
positioning,
quotes
}
\usetikzlibrary{matrix,calc,arrows}
\usetikzlibrary{arrows}
\usepackage{tikz-cd}
\usepackage{caption}
\usepackage{bbm}
\usepackage{amssymb}
\usepackage{graphicx}
\usetikzlibrary{arrows}

\pgfdeclarelayer{nodelayer}
\pgfdeclarelayer{edgelayer}
\pgfdeclarelayer{bg}
\pgfdeclarelayer{background}
\pgfsetlayers{bg,background,edgelayer,nodelayer,main}
\SetVertexStyle[FillColor=white,FillOpacity=0.7,LineOpacity=0.7,MinSize=0.5cm]

\tikzstyle{feature}=[fill=white, 
draw=black, 
shape=circle,
minimum size=1cm]

\tikzstyle{mbfeature}=[fill=lightgray, 
draw=black, 
shape=circle,
minimum size=1cm]           

\tikzstyle{chosen_feature}=[fill=lightgray, 
draw=black, 
shape=circle,
minimum size=1cm]   

\tikzstyle{generator_block}=[fill=white, 
draw=black, 
shape=rectangle]

\tikzstyle{text_only}=[fill=white, 
draw=none, 
shape=rectangle]

\tikzstyle{latent}=[fill=white, 
draw=black,thick,dashed, 
shape=circle,
minimum size=1cm]

\tikzstyle{concat_mb_variables}=[fill=white, 
draw=black, 
shape=rectangle,
minimum size=1cm,
minimum width=2.5cm,
rounded corners=10]
\tikzstyle{none}=[draw=none]
\tikzstyle{noiseterm}=[draw=none]

\tikzstyle{synapse}=[draw=black, -]
\tikzstyle{arrow}=[->, line width=0.2mm]
\tikzstyle{arrow_com}=[<->, line width=0.3mm]
\tikzstyle{arrow_dotted}=[<->, line width=0.3mm,dotted]
\tikzstyle{arrival_departure}=[{|->}, draw={rgb,255: red,52; green,129; blue,230}]
\tikzstyle{part_traj}=[line width=0.2mm, draw={rgb,255: red,66; green,126; blue,255}, ->]
\tikzstyle{death}=[line width=0.2mm, draw=red,->]

\usetikzlibrary{shapes,decorations,arrows,calc,arrows.meta,fit,positioning}
\tikzset{
-Latex,auto,node distance =1 cm and 1 cm,semithick,
state/.style ={circle, draw, minimum width = 0.7 cm},
point/.style = {circle, draw, inner sep=0.04cm,fill,node contents={}},
bidirected/.style={Latex-Latex,dashed},
el/.style = {inner sep=2pt, align=left, sloped}
}

\usetikzlibrary{shapes.geometric, arrows}
\tikzstyle{startstop} = [rectangle, rounded corners, minimum width=3cm, minimum height=1cm,text centered, draw=black]
\tikzstyle{io} = [trapezium, trapezium left angle=70, trapezium right angle=110, minimum width=3cm, minimum height=1cm, text centered, draw=black, fill=blue!30]
\tikzstyle{process} = [rectangle, minimum width=3cm, minimum height=1cm, text centered, draw=black, fill=orange!30]
\tikzstyle{decision} = [diamond, minimum width=3cm, minimum height=1cm, text centered, draw=black, fill=green!30]

\usepackage{algorithm}
\usepackage{algpseudocode}
\usepackage{mathtools}

\DeclareMathOperator*{\argmax}{arg\,max}
\DeclareMathOperator*{\argmin}{arg\,min}
\newcommand\independent{\protect\mathpalette{\protect\independenT}{\perp}}
\def\independenT#1#2{\mathrel{\rlap{$#1#2$}\mkern2mu{#1#2}}}

\newcommand{\parent}[1]{\textbf{Pa}_{#1}}

\newcommand{\xmb}[0]{X_\textbf{MB}}
\newcommand{\muchless}[0]{\texttt{<{}<}}
\newcommand\matchdist{\stackrel{\mathclap{\scriptsize\mbox{d}}}{=}}
\newcommand\tdots{\hbox to 1em{.\hss.\hss.}}

\newcommand\blfootnote[1]{
    \begingroup
    \renewcommand\thefootnote{}\footnote{#1}
    \addtocounter{footnote}{-1}
    \endgroup
}

\SetCommentSty{mycommfont}

\usepackage{fancyhdr}
\usepackage{hyperref}   
\usepackage{ragged2e}   

\fancypagestyle{titlepagefooter}{%
  \fancyhf{} 

  \fancyfoot[C]{%
    \footnotesize
    \parbox{\textwidth}{%
      \justifying
      \hspace{0.05em}
      \\[3pt]
      \noindent©~2025 IEEE. Personal use of this material is permitted. Permission from IEEE must be obtained for all other uses, in any current or future media, including reprinting/republishing this material for advertising or promotional purposes, creating new collective works, for resale or redistribution to servers or lists, or reuse of any copyrighted component of this work in other works. This is the author’s accepted manuscript of the paper published in \textit{IEEE Transactions on Neural Networks and Learning Systems}. The final published version is available at
      \href{https://doi.org/10.1109/TNNLS.2024.3392750}{https://doi.org/10.1109/TNNLS.2024.3392750}.
    }%
  }%
}

\begin{document}

\title{Semi-Supervised Learning under General Causal Models}
\author{Archer Moore,~Heejung Shim,~Jingge Zhu,~Mingming Gong}
\maketitle

\thispagestyle{titlepagefooter}

\definecolor{CGAN_GUMBEL_SUPERVISED_CLASSIFIER}{HTML}{00CC96}
\definecolor{VAT}{HTML}{AB63FA}
\definecolor{ENTROPY_MINIMISATION}{HTML}{B6E880}
\definecolor{SSL_GAN}{HTML}{19D3F3}
\definecolor{CGAN_BASIC_SUPERVISED_CLASSIFIER}{HTML}{EF553B}
\definecolor{SSL_VAE}{HTML}{FF6692}
\definecolor{FULLY_SUPERVISED_CLASSIFIER}{HTML}{034363}
\definecolor{TRIPLE_GAN}{HTML}{FFA15A}
\definecolor{PARTIAL_SUPERVISED_CLASSIFIER}{HTML}{FF97FF}
\definecolor{LABEL_PROPAGATION}{HTML}{873E23}
\definecolor{ASSFSCMR}{HTML}{FECB52}
\definecolor{SFAMCAMT}{HTML}{24F239}

\renewcommand{\listofalgorithms}


{\noindent\blfootnote{%
Archer Moore, Heejung Shim, and Mingming Gong are with the School of Mathematics and Statistics, Faculty of Science, The University of Melbourne, Melbourne, Australia (e-mail: apmoore@student.unimelb.edu.au; hee.shim@unimelb.edu.au; mingming.gong@unimelb.edu.au). Jingge Zhu is with the Department of Electrical and Electronic Engineering, Faculty of Engineering and Information Technology, The University of Melbourne (email: jingge.zhu@unimelb.edu.au).%
}}
\setlength{\parindent}{0pt}

\begin{abstract}
Semi-supervised learning (SSL) aims to train a machine learning model using both labelled and unlabelled data. While the unlabelled data have been used in various ways to improve the prediction accuracy, the reason why unlabelled data could help is not fully understood. One interesting and promising direction is to understand SSL from a causal perspective. In light of the independent causal mechanisms principle, the unlabelled data can be helpful when the label causes the features but not vice versa. However, the causal relations between the features and labels can be complex in real world applications. In this paper, we propose a SSL framework that works with general causal models in which the variables have flexible causal relations. More specifically, we explore the causal graph structures and design corresponding causal generative models which can be learned with the help of unlabelled data. The learned causal generative model can generate synthetic labelled data for training a more accurate predictive model. We verify the effectiveness of our proposed method by empirical studies on both simulated and real data. 
\end{abstract}

\section{Introduction}

The goal of classification is to identify a meaningful attribute from contextual information. This encompasses a myriad of useful research applications, including the identification of pedestrians for self-driving cars \cite{computer_vision_state_of_art_2017} or misinformation in social media networks \cite{identifying_spam_2020}. While modern technology engenders large datasets, unsolved machine learning (ML) classification problems require human identification of ground-truth labels. This may be slow and costly. Such issues motivate Semi-Supervised Learning (SSL), where we aim to use small number of ground-truth labelled examples, as well as a larger set of unlabelled examples, to achieve the ultimate objective of class prediction. One useful application of SSL is to develop medical diagnostic models which currently necessitate finite expert knowledge \cite{semisup_medical,deep_reap_ssl_multipathology}. In a different context, one recent work employs SSL methods for the detection of fake online review data \cite{fake_review_ssl_PU_2023}.

Given some label $Y$, with observable feature variables $X$, the goal of semi-supervised classification is to estimate $P(Y|X)$ from both labelled and unlabelled data \cite{deeplearning_book}. A broad range of approaches for semi-supervised learning exist \cite{deep_ssl_survey,an_overview_of_deep_ssl_2020}, encompassing discriminative and generative methods.
Discriminative approaches use only the unlabelled empirical sample $P(X)$ with regularisation heuristics which link $P(X)$ to $P(Y|X)$, and avoid modelling $P(X)$ directly \cite{deeplearning_book}. Such models are more parsimonious and require fewer assumptions. In contrast, generative approaches model the distribution $P(X)$ or $P(X|Y)$ with a generator $G$, using synthetic samples drawn from $G$ to improve the discriminator. The rationale follows from a rewriting of the classification probability according to Bayes rule:
$P(Y|X)={P(X,Y)}/{P(X)}$.
It is worth mentioning that there are a large number of methods specifically designed for image data: for example, by using consistency regularisation under different augmentations of the input images \cite{an_overview_of_deep_ssl_2020}. 

Despite the success of SSL, the reason why unlabelled data could help is still not fully understood. An interesting direction of SSL research is to understand SSL from a causal perspective. In the seminal work \cite{anticausal_learning}, the authors consider two learning scenarios under two possible causal graphs between labels and features, including causal learning $X\rightarrow Y$ and anticausal learning $Y\rightarrow X$, and conjecture that SSL methods can only be successful in the anticausal scenario with empirical illustrations. This conjecture is proven from an information-theoretical perspective under a class of parametric models in a recent work \cite{causal_ssl_parametric}.  \cite{vank_ssl} considers the scenario where part of the features are causes of the label and the other part is the effect of label and propose two more general SSL approaches based on generative modeling and self-training, respectively. While outcomes along this line are encouraging, the existing methods are limited to relatively simple causal graphs and cannot handle real situations well.

In this paper, we propose a semi-supervised learning framework under general causal models. More specifically, we consider the possible causal graphs one can obtain from real-word applications and divide them into different categories. Under each category, we develop semi-supervised approaches based on causal generative models, i.e., generative models following causal structures, to effectively leverage labelled and unlabelled data for training predictive models. To the best of our knowledge, this is the first comprehensive causal SSL solution that is exhaustive in this respect. In comparative experiments, our method demonstrates superior performance over all models on synthetic and real-world data. We share code used in our experiments \footnote{\url{https://github.com/apmoore499/SSL_GCM}}.



\section{Related Works}
There is a vast literature on semi-supervised learning. Broadly speaking, we can divide existing SSL methods into three categories: discriminative methods, generative methods, and hybrid methods.

{\bf Discriminative methods} train a predictive model by minimizing the risk on labelled data and adding regularisations built from unlabelled data. Entropy minimisation (ENT-MIN) \cite{entmin_orig} optimises the decision boundary to lie within a low-density region, and is grounded in the assumption that features in the same cluster should belong to the same class. A more elaborate recent contribution uses an entropy-based clustering approach in the latent space to improve the quality of representations \cite{ssl_feature_learn_deep_entropy_sparsity_2022}. Pseudo-labelling (also known as label propagation) \cite{pseudolab_orig} uses the softmax label estimates of unlabelled data as a regularisation technique. At each training epoch, softmax predictions are compared with hard labels. The optimisation objective encourages the classifier to make predictions that are closer to the hard labels. Meta pseudolabelling (MPL) \cite{meta_pseudo_label} is an extension which uses a Student-Teacher architecture. Student-Teacher architectures consist of a Teacher discriminator which learns the decision function $P(Y|X)$, and a Student model which learns to emulate the decision predictions of the Teacher. Some further methods are Co-Training \cite{co_training}, which trains on complementary aspects of the data, and TriNET\cite{trinet_device}, which forms consensus among three pseudo-labelling models. In general, these methods suffer from confirmation-bias, as models are incentivised to produce consistent predictions over successive training epochs, and minimise uncertainty in ambiguous cases \cite{confirmation_bias_ssl_2019}. If the features are multimodal and exist in a number of different domains such as text, images, or video, some interesting works in this area propose to regularise the representation by trying to optimise a lower-dimensional manifold \cite{ssfa_cor_2016}, or by clustering observations in latent space according to their class label \cite{adapt_ssfs_2018}.

Consistency regularisation methods assume that predictions on unlabeled data should be consistent over different discriminative models or a model on different augmentations of input $X$. Virtual adversarial training (VAT) \cite{VAT_orig} extends the adversarial framework of the generative adversarial network (GAN). For a perturbation $\gamma$ in the most vulnerable direction within a small radius around a data point $x\in X$, the model encourages identical predictions for $P(Y|x)$ and $P(Y|x+\gamma)$. Given that the ground-truth labels are never observed for unlabelled data, the loss is denoted as `virtual'. Further approaches abound: Virtual adversarial dropout (VAdD) \cite{adv_dropout_2018} perturbs network weights and changes network structure to increase consistency of predictions that are vulnerable to adversarial attack. Stochastic Weight Averaging \cite{swa_2018} considers various schemes to average weights at differing points during training. Unsupervised data augmentation (UDA) is a more exhaustive approach to text and image augmentation \cite{UWA_2020}. More complicated methods such as the $\Pi$ model \cite{PI_model} enforce consistency regularisation across differing augmentations of identical data, dropout in neural network layers, or random max pooling. There exist an extensive suite of approaches which enforce consistencies across multiple networks: Dual Student \cite{dual_student}, Temporal Ensembling \cite{temporal_ensembling} and Mean Teacher \cite{Mean_Teacher_2017}.

{\bf Generative methods} typically model the marginal distribution $P(X)$ by a generative model and make use of it to improve the prediction performance. Among the generative methods, GAN architectures \cite{GANs_2014} have shown considerable success in SSL. The adversarial framework employs a discriminator $D:\mathcal{X} \rightarrow d, 0\leq d \leq 1$, which is a judge of how accurately G models $P(X)$. Semi-supervised GAN (SGAN)/Improved GAN is the most direct implementation for SSL, using $D$ as both a judge of $G$, and an auxiliary classifier over $k$ classes \cite{ssl_gan_original}. Fake data is mapped to some extra hypothetical class $k+1$. Triple GAN is an alteration to SGAN which separates the discriminator into a real/fake discriminator and an auxiliary classifier for the class \cite{triplegan_orig}. There are many SSL models employing simliar frameworks: Good semi-supervised learning that requires a bad GAN (GoodBadGAN) \cite{neurips_goodbadgan_2017}, localised GAN \cite{local_GAN_2018}, bidirectional GAN (BiGAN) \cite{BiGAN_2017}, Triangle GAN \cite{triangle_GAN_2017}, Structured GAN \cite{structured_GAN_2017}, and others \cite{an_overview_of_deep_ssl_2020,deep_ssl_survey}. These extensions exploit specific assumptions of the data generating process, such as semantic segmentation, local feature maps, latent feature maps, or domain shift: these may be employed where appropriate for specific datasets. 

A different type of generative model implements the Variational Autoencoder (VAE) architecture for use in SSL \cite{kingma2014semisupervised,kingma2014autoencoding}. As per regular VAE, the observational distribution is a function of a set of low-dimensional, independent, normally-distributed latent factors $Z$. VAE methods exploit an identical parameterisation of the latent space for both labelled and unlabelled data. In their original paper, the authors explored a variety of architectures: the most effective model, which we refer to as SSVAE (commonly referred to as M1+M2 in the literature), combines two VAE networks M1 and M2. M1 learns a representation of latent variables $Z_1$ of the marginal distribution $P(X)$ only. M2 then learns a representation of some latent $Z_2$ which depends on the both the label $Y$ and $Z_1$. The marginal probability $P(Z_2|Z_1)$ is used to infer $P(Y|X)$ for unlabelled $X$. Reparameterised VAE (ReVAE)\cite{rethinking_ssl_vae} and auxiliary deep generative models (ADGM) \cite{ADGModels} extend SSVAE by introducing auxiliary variables. Infinite VAE \cite{infinite_VAE} combines multiple individual VAE models. Disentangled VAE \cite{disentangled_VAE} and semi-supervised sequential variational autoencoder (SVAE) \cite{SVAE} enforce stricter assumptions on the data structure. In comparison to SSVAE (M1+M2) and ADGM, semi-supervised Gaussian mixture autoencoder (SeGMA) \cite{ssl_segma_autoencoder_2021} is a recent autoencoder approach which demonstrates superior consistency under style mixing and interpolation between classes.

The existing generative models either directly model $P(X)$ or $P(Y|X)$, which implicitly assume that the underlying causal generative process follows $Y\rightarrow X$. While these methods work well for image data, they do not work well if the underlying causal model is not $Y\rightarrow X$. Therefore, in the current work we wish to explore the effect of a causal structure on the performance of these models, and consider learning under general causal generative models.

{\bf Hybrid Methods} combine elements of methods previously discussed. A central feature of most of these approaches is Mixup \cite{mixup_orig_2017}, which interpolates synthetic samples from labelled pairs indexed by a convex combination of both features and labels. ICT \cite{ICT_2019} implements Mixup consistency regularisation in a student-teacher framework. Recently, interpolation-based contrastive learning for few-label SSL (ICL-SSL) \cite{interpolation_CL_fewlabel_SSL_2022} explores an interpolation scheme which preserves semantic consistency. Mixup has also been combined with meta-learning for considerable SSL performance \cite{meta_mixup_2022}. Mixmatch \cite{mixmatch}, ReMixMatch \cite{berthelot2019remixmatch} and Fixmatch\cite{fixmatch_2020} comprise a similar vein of work combining elements of data augmentation, consistency regularisation and/or pseudo-labelling. These particular methods are applicable to image data, and are not used for comparison in our current work. Broadly, these models find success in implementing various aspects of consistency regularisation to maximise discriminative confidence, while also incorporating assumptions about the data generating process.







\section{Problem setup and notation}
Let two random variables $X$ and $Y$ denote the features and the labels, respectively. In semi-supervised learning, we have labelled examples $\{(x_1,y_1),\dots,(x_l,y_l)\}\in\mathcal{X}\times\mathcal{Y}$ drawn from an unknown joint distribution $P(X,Y)$ and unlabelled examples $\{x_{l+1},\dots,x_{l+u}\}\in\mathcal{X}$ drawn from the marginal distribution $P(X)$. The labelled examples are usually much less than unlabelled examples, i.e., $l\muchless u$. The goal of semi-supervised learning is to learn a predictive function $f$ from both labelled and unlabelled data, with the expectation that $f$ performs better than the function learned from only labelled data. In this paper, we consider the classification problem with $\mathcal{Y}=\{1,\ldots,K\}$ and $\mathcal{X}=\mathbb{R}^d$.

\section{Preliminaries}

\subsection{Causal Models}

$\noindent\textbf{Directed Acyclic Graph (DAG)}$. We require a framework for articulating causal relationships between all variables in $V=\{X,Y\}$. Following Pearl's framework \cite{inference_pearl_systems}, we use the Directed Acyclic Graph (DAG) to represent causal interactions. A DAG is a directed graph (DiGraph) which is acyclic. A DiGraph $\mathcal{G}=\langle\mathcal{V},\mathcal{E}\rangle$ is a pair over vertices $\mathcal{V}$ and edges $\mathcal{E}$, where each edge $e\in\mathcal{E}$ is an ordered tuple $(v_i,v_k)$ over two distinct vertices $v_i,v_k\in \mathcal{V}$. 
The direction is visually depicted using an arrowhead. A path is an ordered collection of edges which share common source/target vertices. If the DiGraph contains edges $e_x=(v_i,v_k),e_{x'}=(v_k,v_m)$, we say that there is a path from $v_i$ to $v_m$. A DiGraph is acyclic if there exist no paths which start and end at the same vertex. If there is an edge from $v_i$ to $v_k$, we say that $v_i$ is a parent of $v_k$. We denote all parents of $v_k$ as $\parent{v_k}$. If $v_k$ has no parents, then $\parent{v_k}=\{\emptyset\}$ and $v_k$ is called a root node. Considering the correspondence between vertices $v_i\in\mathcal{V}$ and variables in $V=\{X,Y\}$, we use $v_i$ to interchangeably refer to a variable in $V$, or its corresponding vertex in $\mathcal{G}$. 

The DAG can be used as a representation of causal mechanisms between all variables $v_i\in\{X,Y\}$, and the causal mechanism of $v_i$ is represented by the expression $P(v_i|\parent{v_i})$. Using a causal framework, we seek an understanding of how the observational distribution $P(X,Y)$ may arise from a hierarchy of causal mechanisms. In addition to assuming a DAG representation, we require further assumptions on the data generating process: these are encompassed by causal Bayesian networks and structural causal models (SCMs).

$\noindent\textbf{Causal Bayesian Networks}$. For some DAG $\mathcal{G}=\langle\mathcal{V},\mathcal{E}\rangle$, let each vertex $v_i\in \mathcal{V}$ correspond to a particular variable $v_i\in V=\{X,Y\}$ and let $P$ be a joint probability distribution over $V$. The pair $\langle \mathcal{G},P \rangle$ is a causal Bayesian network if the Causal Markov condition and Modularity condition  hold. The $\textbf{Causal Markov condition}$ implies that the joint distribution $P(V)$ can be factorised into the product of each $v_i\in V$ conditional on its parents $\parent{v_i}$:
\begin{equation} \label{eq:causal_bayesian_network}
P(V)=\prod\limits_{i=1}^{d} P(v_i|\parent{v_i}).
\end{equation}
The DAG entails that if $\parent{v_i}$ is a cause of $v_i$, then an intervention on $\parent{v_i}$ should affect the value of $v_i$. In contrast, an intervention on $v_i$ does not imply a change in $\parent{v_i}$. This is the $\textit{Modularity principle}$ \cite{inference_pearl_systems}: the causal mechanism of unintervened variables is invariant under intervention. While causal Bayesian networks are an elegant representation of a joint probability, they may suffer from identifiabiliity guarantees as, in general, it can be difficult to confirm or deny a unique causal DAG $\mathcal{G}$ for some $P$. In this project, we avoid this limitation by assuming a prior knowledge of the DAG. 

\textbf{Structural Causal Model.} SCM is an alternative way to formalise a model of the causal mechanism for each $v_i\in V$. The SCM consists of a tuple $\langle S,P \rangle$ where $S=(S_1,\dots,S_d)$:
\begin{equation} \label{eq:scm} 
S_i:= f_i(\parent{X_i}, N_i) \hspace{1cm} i=1,...,d,
\end{equation} 
where each $S_i$ corresponds to each $v_i\in\{V\}$. $N_i$ is an approximation of all possible external causes of $v_i$, including errors in measurement. Writing $N=\{N_1,\dots,N_d\}$, all $N_i$ are mutually independent:
\begin{equation}
P(N)=\prod\limits_{i=1}^{d} P(N_i).
\end{equation}
In the SCM framework, $f_i$ describes a deterministic relationship between $v_i$ and its parents. This is grounded in the assumption that the causal mechanism is an interaction of physical phenomena, and is therefore encompassed by some set of deterministic set of physical laws \cite{inference_pearl_systems}. $f_i$ is a function of the random variable $N_i$, implying that $S_i$ is also a random variable. However, $f_i$ itself does not change, as the laws of physics are not mutable. This is known as the $\textit{Independent Causal Mechanisms}$ (ICM) principle. The ensemble of causal mechanisms described by each $S_i$ is a model of $P(v_i|\parent{v_i})$, and the hierarchical collection of all $S_i$ describes the generative process of how i.i.d. samples of $P(X,Y)$ arise. 




\subsection{Markov Blanket}

While previous discussion illustrated the relevance of causality for modelling $P(X,Y)$, the causal DAG can also inform which features are relevant for estimating $P(Y|X)$. The classification task can be simplified by identifying and using features in the Markov Blanket of $Y$, written as $\xmb$. The Markov Blanket of $Y$ is the set of features $\xmb\in X \text{ s.t. } Y \independent{X\setminus \xmb} | \xmb$. Intuitively, this means that given information contained in the features in $\xmb$, $Y$ is conditionally independent of all other features. Considering our classification objective is to estimate $P(Y|X)$, the following equivalence should therefore hold:
\begin{equation} \label{eq:mb_condition}
    P(Y|X)=P(Y|\xmb),
\end{equation}
where $\xmb$ contains features $X$ which are either parents $X_C$, spouses $X_S$ or children $X_E$ of $Y$. The associated causal relationships are depicted in Table \ref{tab:parent_spouse_effect_terminology}.

\begin{table}[H]
\centering
    \begingroup
    \setlength{\tabcolsep}{6pt} 
    \renewcommand{\arraystretch}{2} 
\centering
\begin{tabular}{|M{3.2cm}|M{2cm}|M{3.5cm}|}
\hline
\textbf{Terminology}          & \textbf{Notation} & \textbf{Example}                          \\ \hline
Direct cause / parent of $Y$ & $X_C$    & \scalebox{0.6}{\begin{tikzpicture}
	\begin{pgfonlayer}{nodelayer}
		\node [style=feature] (0) at (-3.5, 0) {{$X_C$}};
		\node [style=feature] (1) at (-1, 0) {{$Y$}};
	\end{pgfonlayer}
	\begin{pgfonlayer}{edgelayer}
		\draw [-latex] (0) to (1);
	\end{pgfonlayer}
\end{tikzpicture}} \\ \hline
Direct effect / child of $Y$ & $X_E$    & \scalebox{0.6}{\begin{tikzpicture}
	\begin{pgfonlayer}{nodelayer}
		\node [style=feature] (0) at (-3.5, 0) {{$Y$}};
		\node [style=feature] (1) at (-1, 0) {{$X_E$}};
	\end{pgfonlayer}
	\begin{pgfonlayer}{edgelayer}
		\draw [-latex] (0) to (1);
	\end{pgfonlayer}
\end{tikzpicture}} \\ \hline
Spouse of $Y$               & $X_S$    & \scalebox{0.6}{\begin{tikzpicture}
	\begin{pgfonlayer}{nodelayer}
		\node [style=feature] (xs) at (-3.5, 0) {{$Y$}};
		\node [style=feature] (xe) at (-1, 0) {{$X_E$}};
		\node [style=feature] (y) at (1.5, 0) {{$X_S$}};
	\end{pgfonlayer}
	\begin{pgfonlayer}{edgelayer}
		\draw [-latex] (xs) to (xe);
		\draw [-latex] (y) to (xe);
	\end{pgfonlayer}
\end{tikzpicture}} \\ \hline
\end{tabular}
\endgroup
\captionsetup{justification=centering,width=1.0\textwidth}
\caption{Parents, children and spouses in $\xmb$}
\label{tab:parent_spouse_effect_terminology}
\end{table}

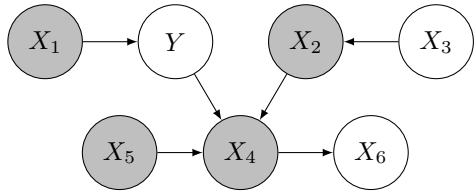
\begin{figure}[H]
\centering
\begin{adjustbox}{minipage=\linewidth,scale=0.98}
\centering
\begin{tikzpicture}
	\begin{pgfonlayer}{nodelayer}
		\node [style=mbfeature] (x1) at (-3.5, 0.5) {{$X_1$}};
		\node [style=feature] (y) at (-1.75, 0.5) {{$Y$}};
		\node [style=mbfeature] (x2) at (0, 0.5) {{$X_2$}};
		\node [style=feature] (x3) at (1.75, 0.5) {{$X_3$}};
		\node [style=mbfeature] (x4) at (-0.875, -1.0) {{$X_4$}};
		\node [style=mbfeature] (x5) at (-2.5, -1.0) {{$X_5$}};
		\node [style=feature] (x6) at (0.875, -1.0) {{$X_6$}};

	\end{pgfonlayer}
	\begin{pgfonlayer}{edgelayer}
		\draw [-latex] (x1) to (y);
		\draw [-latex] (y) to (x4);
		\draw [-latex] (x2) to (x4);
        \draw [-latex] (x5) to (x4);
        \draw [-latex] (x3) to (x2);
        \draw [-latex] (x4) to (x6);
	\end{pgfonlayer}
\end{tikzpicture}
\captionsetup{justification=centering,width=0.99\textwidth}
\caption{Markov Blanket $\xmb=\{X_1,X_5,X_4,X_2\}$}
\label{fig:mb_illustration}
\end{adjustbox}
\end{figure}

In Figure \ref{fig:mb_illustration}, we illustrate an example where features in $\xmb$ are shaded grey, and $\xmb=\{X_1,X_5,X_4,X_2\}$, since $X_1$ is a direct cause (parent) of $Y$, $X_4$ is a direct effect (child) of $Y$ and $X_2,X_5$ are both spouses of $Y$. In contrast, $X_3$ and $X_6$ are neither parents, spouses or children of $Y$, and hence are not in $\xmb$. For a classification task, observations of $X_1,X_2,X_4,X_5$ are sufficient for estimating $P(Y|X)$: we gain no extra information from $X_3,X_6$ when all features $X\in\xmb$ are observed.

\section{Causal Semi-Supervised Learning using $\xmb$}

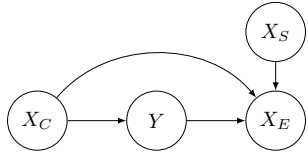
\begin{figure}[H]
\begin{adjustbox}{minipage=\linewidth,scale=0.8,center}
\centering
\scalebox{0.98}{\begin{tikzpicture}

	\begin{pgfonlayer}{nodelayer}
	\node [style=feature] (xc) at (-2, 0) {{$X_C$}};
	\node [style=feature] (y) at (0, 0) {{$Y$}};
	\node [style=feature] (xe) at (2, 0) {{$X_{E}$}};
	\node [style=feature] (xs) at (2, 1.5) {{$X_{S}$}};
	\end{pgfonlayer}

	\begin{pgfonlayer}{edgelayer}
    \draw [-latex] (xc) to (y);
    \draw [-latex] (y) to (xe);
    \draw [-latex] (xs) to (xe);
	\draw[-latex] (xc) to[out=50,in=130] (xe);
	\end{pgfonlayer}

\end{tikzpicture}}
\captionsetup{justification=centering,width=1\textwidth}
\caption{CG6: Markov Blanket containing $X_S,X_C,X_E$}
\label{fig:markov_blanket_full}
\end{adjustbox}
\end{figure}

Our goal is to exploit the causal structure underlying $P(X,Y)$ for semi-supervised learning. This task is simplified considering Equation \ref{eq:mb_condition}, as the Markov Blanket contains all useful information for estimating $P(Y|X)$. In the current section, we analyse the plausibility of SSL over any Markov Blanket structure. Such results extend naturally to any DAG, providing a unified framework for causal SSL. A key conjecture of previous works is that for data $P(X,Y)$, unlabelled samples could facilitate SSL if the causal structure is $Y\rightarrow X$, but not if $X\rightarrow Y$\cite{anticausal_learning,info_geo_infer_causal_dir,element_of_ci_book_2017}, with the relationship proven analytically for a class of parametric models\cite{causal_ssl_parametric}. Since such results account for Markov Blankets containing either $X_C$ or $X_E$ only, we expand the analysis to consider how more features could be used. To gain some intuition for different types of Markov Blanket structures, consider the DAG in Figure \ref{fig:markov_blanket_full}, where $X_S,X_C,X_E\in\xmb$.

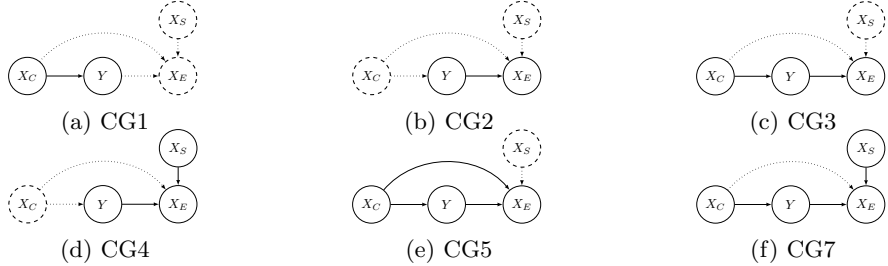
\begin{figure}[H]
\centering
\begin{adjustbox}{minipage=\linewidth,scale=0.99,left,lap=0.01\width}
\centering
\begin{subfigure}[t]{0.3\textwidth}
\centering
\scalebox{0.5}{\begin{tikzpicture}

	\begin{pgfonlayer}{nodelayer}
	\node [style=feature] (xc) at (-2, 0) {{$X_C$}};
	\node [style=feature] (y) at (0, 0) {{$Y$}};
	\node [style=latent] (xe) at (2, 0) {{$X_{E}$}};
	\node [style=latent] (xs) at (2, 1.5) {{$X_{S}$}};
	\end{pgfonlayer}

	\begin{pgfonlayer}{edgelayer}
    \draw [-latex] (xc) to (y);
    \draw [-latex,dotted] (y) to (xe);
    \draw [-latex,dotted] (xs) to (xe);
	\draw[-latex,dotted] (xc) to[out=50,in=130] (xe);
	\end{pgfonlayer}

\end{tikzpicture}}
\captionsetup{justification=centering,width=0.6\textwidth}
\caption{CG1}
\end{subfigure}%
\begin{subfigure}[t]{0.3\textwidth}
\centering
\scalebox{0.5}{\begin{tikzpicture}

	\begin{pgfonlayer}{nodelayer}
	\node [style=latent] (xc) at (-2, 0) {{$X_C$}};
	\node [style=feature] (y) at (0, 0) {{$Y$}};
	\node [style=feature] (xe) at (2, 0) {{$X_{E}$}};
	\node [style=latent] (xs) at (2, 1.5) {{$X_{S}$}};
	\end{pgfonlayer}

	\begin{pgfonlayer}{edgelayer}
    \draw [-latex,dotted] (xc) to (y);
    \draw [-latex] (y) to (xe);
    \draw [-latex,dotted] (xs) to (xe);
	\draw[-latex,dotted] (xc) to[out=50,in=130] (xe);
	\end{pgfonlayer}

\end{tikzpicture}}
\captionsetup{justification=centering,width=0.6\textwidth}
\caption{CG2}
\end{subfigure}%
\begin{subfigure}[t]{0.3\textwidth}
\centering
\scalebox{0.5}{\begin{tikzpicture}

	\begin{pgfonlayer}{nodelayer}
	\node [style=feature] (xc) at (-2, 0) {{$X_C$}};
	\node [style=feature] (y) at (0, 0) {{$Y$}};
	\node [style=feature] (xe) at (2, 0) {{$X_{E}$}};
	\node [style=latent] (xs) at (2, 1.5) {{$X_{S}$}};
	\end{pgfonlayer}

	\begin{pgfonlayer}{edgelayer}
    \draw [-latex] (xc) to (y);
    \draw [-latex] (y) to (xe);
    \draw [-latex,dotted] (xs) to (xe);
	\draw[-latex,dotted] (xc) to[out=50,in=130] (xe);
	\end{pgfonlayer}

\end{tikzpicture}}
\captionsetup{justification=centering,width=0.6\textwidth}
\caption{CG3}
\end{subfigure}\vspace{-1em}\\
\begin{subfigure}[t]{0.3\textwidth}
\centering
\scalebox{0.5}{\begin{tikzpicture}

	\begin{pgfonlayer}{nodelayer}
	\node [style=latent] (xc) at (-2, 0) {{$X_C$}};
	\node [style=feature] (y) at (0, 0) {{$Y$}};
	\node [style=feature] (xe) at (2, 0) {{$X_{E}$}};
	\node [style=feature] (xs) at (2, 1.5) {{$X_{S}$}};
	\end{pgfonlayer}

	\begin{pgfonlayer}{edgelayer}
    \draw [-latex,dotted] (xc) to (y);
    \draw [-latex] (y) to (xe);
    \draw [-latex] (xs) to (xe);
	\draw[-latex,dotted] (xc) to[out=50,in=130] (xe);
	\end{pgfonlayer}

\end{tikzpicture}}
\captionsetup{justification=centering,width=0.6\textwidth}
\caption{CG4}
\end{subfigure}%
\begin{subfigure}[t]{0.3\textwidth}
\centering
\scalebox{0.5}{\begin{tikzpicture}

	\begin{pgfonlayer}{nodelayer}
	\node [style=feature] (xc) at (-2, 0) {{$X_C$}};
	\node [style=feature] (y) at (0, 0) {{$Y$}};
	\node [style=feature] (xe) at (2, 0) {{$X_{E}$}};
	\node [style=latent] (xs) at (2, 1.5) {{$X_{S}$}};
	\end{pgfonlayer}

	\begin{pgfonlayer}{edgelayer}
    \draw [-latex] (xc) to (y);
    \draw [-latex] (y) to (xe);
    \draw [-latex,dotted] (xs) to (xe);
	\draw[-latex] (xc) to[out=50,in=130] (xe);
	\end{pgfonlayer}

\end{tikzpicture}}
\captionsetup{justification=centering,width=0.6\textwidth}
\caption{CG5}
\end{subfigure}%
\begin{subfigure}[t]{0.3\textwidth}
\centering
\scalebox{0.5}{\begin{tikzpicture}

	\begin{pgfonlayer}{nodelayer}
	\node [style=feature] (xc) at (-2, 0) {{$X_C$}};
	\node [style=feature] (y) at (0, 0) {{$Y$}};
	\node [style=feature] (xe) at (2, 0) {{$X_{E}$}};
	\node [style=feature] (xs) at (2, 1.5) {{$X_{S}$}};
	\end{pgfonlayer}

	\begin{pgfonlayer}{edgelayer}
    \draw [-latex] (xc) to (y);
    \draw [-latex] (y) to (xe);
    \draw [-latex] (xs) to (xe);
	\draw[-latex,dotted] (xc) to[out=50,in=130] (xe);
	\end{pgfonlayer}

\end{tikzpicture}}
\captionsetup{justification=centering,width=0.6\textwidth}
\caption{CG7}
\end{subfigure}\vspace{-1em}\\
\end{adjustbox}
\captionsetup{justification=centering,width=0.8\textwidth}
\caption{Markov Blanket Subgraphs of CG6}
\label{fig:mb_topologies}
\end{figure}

There are 6 subgraphs of Figure \ref{fig:markov_blanket_full} which are also a Markov Blanket over $Y$. Each variation is depicted in Figure \ref{fig:mb_topologies}, with dotted elements excluded. If we group all variations based on whether $X_C,X_E,X_S\in\xmb$, we see that there are five possible cases. Using notation $\xmb^{1},\dots,\xmb^{5}$ for each case, we summarise these in Table \ref{tab:scenario_ssl_mb}. For example, $\xmb^1$ is a Markov Blanket which contains direct causes $X_C$ of $Y$, no spouses $X_S$ and no children $X_E$. Clearly then, CG1 corresponds to $\xmb^1$. In contrast, $\xmb^5$ is a Markov Blanket containing parents $X_C$, children $X_E$ and spouses $X_S$. This topology applies to CG6, CG7, and Figure \ref{fig:mb_illustration}.

\begin{figure}[H]
\centering
\begin{table}[H]
\centering
    \begingroup
    \setlength{\tabcolsep}{7pt} 
    \renewcommand{\arraystretch}{1.5} 
    
    \begin{tabular}{|c|c|c|c|c|}
    \hline
        $\xmb$ & $X_C\in\xmb$ & $X_E\in\xmb$ & $X_S\in\xmb$ & \textbf{DAG} \\ \hline
        $\xmb^1$ & yes & no & no & CG1 \\ \hline 
        $\xmb^2$ & no & yes & no & CG2 \\ \hline 
        $\xmb^3$ & no & yes & yes & CG4  \\ \hline 
        $\xmb^4$ & yes & yes & no & CG3, CG5\\ \hline 
        $\xmb^5$ & yes & yes & yes & CG6, CG7 \\ \hline 
    \end{tabular}
    \endgroup
\captionsetup{justification=centering}
\caption{Possible $\xmb$ topologies}
\label{tab:scenario_ssl_mb}
\end{table}
\end{figure}

Although in general the structure of a given Markov Blanket may be very complicated, any Markov Blanket will correspond to one of the cases as given in Table \ref{tab:scenario_ssl_mb}. Therefore, if we are able to elucidate the utility of unlabelled data in each case, we provide a unified framework for whether unlabelled data could help over any causal structure. The proposed grouping allows us to make arguments based on a causal factorisation over the DAG. As such, each $\xmb^1,\dots,\xmb^5$ should inherit the same theoretical guarantees. We now discuss each case, starting from $\xmb^{1}$.

$\xmb^1$. Recall that we seek to use the unlabelled data distribution, $P(X_C)$, to improve our estimate of $P(Y|X_C)$. $\xmb^1$ corresponds to CG1 in Figure \ref{fig:mb_topologies}, which has been studied in previous works  \cite{anticausal_learning,causal_ssl_parametric}. The causal factorisation of the the joint distribution is $P(X_C,Y)=P(Y|X_C)P(X_C)$. According to the independent causal mechanism,  $P(Y|X_C)$ and $P(X_C)$ do not contain information about each other, i.e., algorithmically independent \cite{janzing2010causal}. As a result, a better estimation of $P(X_C)$ from unlabeled data could not benefit the estimation of $P(Y|X_C)$. That could improve upon $P(Y|X)$.

$\xmb^2$. This scenario is depicted in CG2 of Figure \ref{fig:mb_topologies}, which is a widely studied scenario \cite {anticausal_learning,kingma2014semisupervised} and it is often applied to image data \cite{kingma2014semisupervised,gong2016domain}. The causal factorisation of the data generating process is: $P(X_E,Y)=P(X_E|Y)P(Y)$. $P(X_E|Y)$ and $P(Y)$ are independent causal mechanisms, i.e., they do not contain information about each other. In contrast, in the anticausal direction, $P(X_E)$ and $P(Y|X_E)$ contain information about each other. This is because $P(X_E)$ is associated with both $P(X_E|Y)$ and $P(Y)$, which induces the labelling function $P(Y|X_E)$.

$\xmb^3$. In this Markov Blanket shown in CG4 of Figure \ref{fig:mb_topologies} , the task is to improve $P(Y|X_E,X_S)$ using the unlabelled data in either $X_E$, $X_S$. This case has some familiar structure to $Y\rightarrow X_E$, but with an additional casual variable of $X_S$. One can compute $P(Y|X_E,X_S)$ as follows:
\begin{align*}
    P(Y|X_E,X_S)&=\frac{P(Y,X_E,X_S)}{P(X_E,X_S)}\\&=\frac{P(X_E|Y,X_S) P(Y)P(X_S)}{\int_Y P(X_E|Y,X_S) P(Y)P(X_S)dY}\\&=\frac{P(X_E|Y,X_S) P(Y)}{\int_Y P(X_E|Y,X_S) P(Y)dY}.
\end{align*}
Obviously, the final predictive distribution $P(Y|X_E,X_S)$ and the marginal distribution $P(X_S)$ contain no information about each other. Thus, the unlabeled data of $X_S$ alone is not beneficial for learning the predictive model. By contrast, the conditional distribution $P(X_E|X_S)=\int_Y P(X_E|Y,X_S) P(Y)dY$ contains information about $P(X_E|Y,X_S)$ and $P(Y)$, which induce the predictive distribution $P(Y|X_E,X_S)$. Therefore, semi-supervised learning would be possible by using unlabeled data to better estimate $P(X_E|X_S)$, which could improve the estimation of $P(Y|X_E,X_S)$.

$\xmb^4$. Two topologies are depicted as CG3 and CG5 in Figure \ref{fig:mb_topologies}. CG5 is analysed in \cite{vank_ssl}, which argues that the clusters $P(X_E|X_C)$ should correspond to different parameterisations based on $Y$. Here we generalize the results to make use of the whole data generating process. More specifically, one can write the predictive distribution as
\begin{align*}
    P(Y|X_E,X_C)&=\frac{P(X_E|Y,X_C)P(Y|X_C)}{\int_Y P(X_E|Y,X_C)P(Y|X_C)dY}.
\end{align*}
It can be seen that $P(X_E|X_C)=\int_Y P(X_E|Y,X_C) P(Y|X_C)dY$ contains information about $P(X_E|Y,X_C)$ and $P(Y|X_C)$, which induce the predictive distribution $P(Y|X_E,X_C)$. In our method section, we will present our algorithms that learn $P(X_E|Y,X_C)$ and $P(Y|X_C)$ with the help of unlabeled data.

As a special case of CG5, CG3 removes the confounding path from $X_C$ to $X_E$. CG3 still benefits from a better estimation of $P(X_E|X_C)$ using unlabeled data, but one can also utilize unlabeled data to estimate $P(X_E)$ alone so as to improve the learning of $P(X_E|Y)$ as in CG2, due to the absence of confounding.





$\xmb^5$. In this Markov Blanket, we aim to predict $Y$ using $X_C,X_S$ and $X_E$: hence the objective is to estimate $P(Y|X_E,X_S,X_C)$. We note that all features $X_C,X_E,X_S$ must be used together. If $X_E$ is parameterised by $X_S,Y,X_C$, as depicted in CG6 of Figure \ref{fig:markov_blanket_full}, we can write the predictive distribution as:
\begin{flalign}
    P(Y|X_E,X_S,X_C)&=\frac{P(Y,X_E,X_S,X_C)}{P(X_E,X_S,X_C)}\nonumber\\
    &=\frac{P(X_E|Y,X_S,X_C) P(Y|X_C)}{\int_Y P(X_E|Y,X_S,X_C)P(Y|X_C)dY}.
    \label{eq:mb5}
\end{flalign}
Here, the conditional distribution $P(X_E|X_S,X_C)=\int_Y P(X_E|Y,X_S,X_C)P(Y|X_C)dY$ contains information about $P(Y|X_C)$ and $P(X_E|Y,X_S,X_C)$, which induce the predictive distribution $P(Y|X_E,X_S,X_C)$. Thus, we can use unlabelled data to better estimate $P(X_E|X_S,X_C)$, which could lead to a more accurate estimate of $P(X_E|Y,X_S,X_C)$ and $P(Y|X_C)$. As a result, the predictive distribution $P(Y|X_E,X_S,X_C)$ induced from $P(X_E|Y,X_S,X_C)$ and $P(Y|X_C)$ can be better estimated. 


{CG7 in Figure \ref{fig:mb_topologies} could be considered similarly to CG4, since $P(X_E|X_S)=\int_Y P(X_E|Y,X_S)P(Y)dY$. These terms could be used to induce the predictive distribution $P(Y|X_S,X_E)$ without confounding from $X_C$. Alternatively, we can also use $X_C$ by considering the conditional probability $P(X_E|X_S,X_C)=\int_Y P(X_E|Y,X_S)P(Y|X_C)dY$. Since it contains information about $P(X_E|Y,X_S)$ and $P(Y|X_C)$, this could be used to induce the predictive disribution $P(Y|X_S,X_E,X_C)$. We should expect CG7 to benefit from an improved estimate of $P(X_E|X_S,X_C)$ in the latter case.}

\section{Method for Causal SSL}
\noindent{\textbf{Method Overview}}. While in theory causal relationships may improve a semi-supervised model, it is not obvious how one is supposed to exploit causal information in a practical modelling context. Our goal is to create a method which could be used to benefit from unlabelled data over any causal structure. Recalling that we assume $P(X,Y)=\prod\limits_{i=1}^{d} P(v_i|\parent{v_i})$, we produce a generative model $G:=\hat{P}(X,Y)$ by creating a structural model for each $P(v_i|\parent{v_i})$ separately. We then synthesise extra data $(x',y')\sim G$, which augments the original labelled sample $(x,y)\in D_l$, and the augmented sample is used to train a classifier in a fully-supervised regime. The motivation for a generative approach is to encode the ICM assumption, and thus the flow of causal information, into the generated data. Our method modifies existing nonparametric approaches to SCM modelling \cite{mmd_gan_2017} in order to estimate $G$ under missing labels. 

\noindent{\textbf{Modelling each }$P(v_i|\parent{v_i})$}. By fully factorising $P(X,Y)=\prod P(v_i|\parent{v_i})$ according to the Causal Markov condition in Equation \ref{eq:causal_bayesian_network}, we identify the factors for structural modelling. In CG4, for example, we can see that there are three factors:

\begin{equation*}
P(X,Y)  =  \underbrace{P(X_E|Y,X_S)}_\text{Factor 1}    \underbrace{P(Y)}_\text{Factor 2}    \underbrace{P(X_S)}_\text{Factor 3}.
\end{equation*}

It is impractical to use an identical estimation procedure for each $P(v_i|\parent{v_i})$, since some factors may need to be estimated in the presence of missing label data, and our choice of loss function depends on whether $v_i\in X$ or $v_i\in Y$. In CG4, $P(X_E|Y,X_S)$ must be estimated in the presence of missing labels, but $P(X_S)$ does not suffer from the same issue. Similarly, our model of $P(Y)$ maps to a label, while $P(X_S)$ maps to a feature. Each factor requires a different approach. To determine the appropriate estimation procedure, we identify some structure in $P(v_i|\parent{v_i})$ according to whether ${v_i}$ is a feature, and whether $\parent{v_i}$ contains any features, any labels, or is empty. There are two sets of approaches we will describe: disjoint, and joint.

\noindent{\textbf{Modelling Approach I: Disjoint.}} Under our disjoint approach, we focus on identifying structure in each $P(v_i|\parent{v_i})$ separately. Our set of rules characterises any single $P(v_i|\parent{v_i})$ as belonging to five separate scenarios, \textbf{A},\textbf{B},\textbf{C},\textbf{D},\textbf{E}, listed in Table \ref{tab:disjoint_modelling_table}. 

\begingroup
\renewcommand{\arraystretch}{1.2}
\begin{table}[H]
\centering
\begin{tabular}{|M{2cm}|M{2.2cm}|M{0.5cm}|M{3.6cm}|}
\hline \textbf{Scenario} &  \textbf{ Description}& $v_i$ & $P(v_i|\parent{v_i})$ \\ \hline
A& $Y$ is a root node, $\parent{Y}=\{\varnothing\}$& $Y$ & P(Y) \\ \hline
B& $Y$ is not a root node, $\parent{Y}\neq\{\varnothing\}$ & $Y$ & $P(Y|X_K,\dots)$ \\ \hline
C&$X_I$ is a root node, $\parent{X_I}=\{\varnothing\}$ & $X_I$ & $P(X_I)$ \\ \hline
D&$X_K\in\parent{X_I}$ and $Y\notin\parent{X_I}$ & $X_I$ & $P(X_I|X_K,\tdots)$ \\ \hline
E&$Y\in\parent{X_I}$ & $X_I$ & $P(X_I|Y,\tdots)$  \\ \hline
\end{tabular}
\captionsetup{justification=centering}
\caption{Disjoint modelling approach for $P(v_i|\parent{v_i})$}
\label{tab:disjoint_modelling_table}
\end{table}
\endgroup

To use this table, we could either match the form of $P(v_i|\parent{v_i})$ to an appropriate expression in the fourth column, or use the \textbf{Description} column. For example, if $P(v_i|\parent{v_i})=P(Y)$, then this form corresponds to scenario \textbf{A}. In contrast, $P(v_i|\parent{v_i})=P(Y|X_C)$ corresponds to scenario \textbf{B}. We use notation $P(v_1|v_2,\tdots)$ to indicate that $\parent{v_1}$ must contain $v_2$, and possibly extra variables. $P(X_1|Y)$, $P(X_1|Y,X_2)$ and $P(X_1|Y,X_2,X_3)$ all conform to the pattern $P(X_I|Y,\dots)$, and therefore they are all identified as scenario \textbf{E}: even though the conditional information is different in each case, all factors depend on $Y$. In contrast, $P(X_1)$ and $P(X_1|X_2)$ cannot conform to this pattern, and are identified as scenario $\textbf{C}$ and $\textbf{D}$ respectively. This structural identification process is used for Markov Blankets CG1-CG7 given earlier. For CG1, the causal factorisation $P(X,Y)=P(Y|X_C)P(X_C)$ contains two causal modules $P(Y|X_C)$ and $P(X_C)$. From Table \ref{tab:disjoint_modelling_table}, it is clear that $P(X_C)$ is scenario \textbf{C} and $P(Y|X_C)$ is scenario \textbf{B}. For CG1-CG7, all factorised modules and scenarios are listed in Figure \ref{fig:disjoint_cg1_cg7}.

\begin{figure}[H]
\centering
\begin{adjustbox}{minipage=\linewidth,scale=0.9,center}
\begin{subfigure}[t]{.5\linewidth}
\scalebox{0.98}{
	
\begin{tabular}{lll}
Graph & Module         & Scenario \\ \hline
CG1   & $P(X_C)$       & C        \\
      & $P(Y|X_C)$     & B        \\ \hline
CG2   & $P(Y)$         & A        \\
      & $P(X_E|Y)$     & E        \\ \hline
CG3   & $P(X_C)$       & C        \\
      & $P(Y|X_C)$     & B        \\
      & $P(X_E|Y)$     & E        \\ \hline
CG4   & $P(Y)$         & A        \\
      & $P(X_S)$       & C        \\
      & $P(X_E|Y,X_S)$ & E        \\
      &                &         
\end{tabular}

}
\end{subfigure}%
\begin{subfigure}[t]{.5\linewidth}
\scalebox{0.98}{

\begin{tabular}{lll}
Graph & Module             & Scenario \\ \hline
CG5   & $P(X_C)$           & C        \\
      & $P(Y|X_C)$         & B        \\
      & $P(X_E|Y,X_C)$     & E        \\ \hline
CG6   & $P(X_C)$           & C        \\
      & $P(X_S)$           & C        \\
      & $P(Y|X_C)$         & B        \\
      & $P(X_E|Y,X_C)$     & E        \\ \hline
CG7   & $P(X_C)$           & C        \\
      & $P(X_S)$           & C        \\
      & $P(Y|X_C)$         & B        \\
      & $P(X_E|Y,X_C,X_S)$ & E       
\end{tabular}

}
\end{subfigure}
\end{adjustbox}
\captionsetup{justification=centering,width=1.0\textwidth}
\caption{Disjoint approach: factors and scenarios for CG1-CG7}
\label{fig:disjoint_cg1_cg7}
\end{figure}

\noindent{\textbf{Modelling Approach II: Joint.}} The scenarios given in Table \ref{tab:disjoint_modelling_table} may be used to characterise any $P(v_i|\parent{v_i})$, and hence determine a method for structural modelling of each factor separately. In contrast, in the joint approach, we identify some shared structure between factors, and these factors are modelled together. Specifically, if $\textbf{B}$ and $\textbf{E}$ both occur in the same factorised $P(X,Y)$, our procedure is modified. While the reasons for this will be explained later, for now we are merely focused on identifying the structures. In this special case, the product $P(X_E|Y,\dots)P(Y|X_C,\dots)$ must be present. This scenario is referred to as \textbf{F}, and entails a different method which replaces the methods of \textbf{B} and \textbf{E}. In Table \ref{tab:joint_modelling_table}, we give an updated set of rules: note that scenarios \textbf{A}, \textbf{C}, \textbf{D} are unchanged. For CG7, the structures identified by our disjoint approach are given in Equation \ref{eq:disjointmeth}:
\begin{equation}
    P(X,Y)  =  \underbrace{P(X_E|Y,X_S)}_\text{E}\underbrace{P(Y|X_C)}_\text{B}  \underbrace{P(X_S)}_\text{C} \underbrace{P(X_C)}_\text{C}.
    \label{eq:disjointmeth}
\end{equation}
In contrast, the scenarios identified by the joint approach are depicted in Equation \ref{eq:jointmeth}. This illustrates that the joint identification scheme is trivially different from the disjoint identification scheme trivial in practice, as it amounts to merely recategorising scenarios $\textbf{B}$ and \textbf{E} as scenario \textbf{F}.
\begin{equation}
    P(X,Y)  =  \underbrace{P(X_E|Y,X_S)P(Y|X_C)}_\text{F}  \underbrace{P(X_S)}_\text{C} \underbrace{P(X_C)}_\text{C}.
    \label{eq:jointmeth}
\end{equation}

\begingroup
\renewcommand{\arraystretch}{1.2}
\begin{table}[H]
\centering
\begin{tabular}{|M{2cm}|M{2.2cm}|M{3.6cm}|}
\hline \textbf{Scenario} &  \textbf{ Description} & $P(v_i|\parent{v_i})$ \\ \hline
A& $Y$ is a root node, $\parent{Y}=\{\varnothing\}$ & P(Y) \\ \hline
C&$X_I$ is a root node, $\parent{X_I}=\{\varnothing\}$ & $P(X_I)$ \\ \hline
D&$X_K\in\parent{X_I}$ and $Y\notin\parent{X_I}$ & $P(X_I|X_K,\tdots)$ \\ \hline
F, \textbf{(Special Case)}&$Y\in\parent{X_I}$ and $\parent{Y}\neq\{\varnothing\}$& $P(X_I|Y,\tdots)P(Y|\tdots)$  \\ \hline
\end{tabular}
\captionsetup{justification=centering}
\caption{Joint modelling approach for $P(v_i|\parent{v_i})$}
\label{tab:joint_modelling_table}
\end{table}
\endgroup

We show how this changes earlier classifications for CG1-CG7 in Table \ref{fig:disjoint_cg1_cg7}. CG1, CG2 and CG4 are unchanged by this new rule, since they do not contain the special case. Structures for CG3, CG5, CG6 and CG7 are reidentified in Table \ref{tab:updatedt3}. 

\begin{table}[H]
\centering
\begin{tabular}{lll}
Graph & Module                     & Scenario \\ \hline
CG3   & $P(X_C)$                   & C        \\
      & $P(X_E|Y)P(Y|X_C)$         & F        \\ \hline
CG5   & $P(X_C)$                   & C        \\
      & $P(X_E|Y,X_C)P(Y|X_C)$     & F        \\ \hline
CG6   & $P(X_C)$                   & C        \\
      & $P(X_S)$                   & C        \\
      & $P(X_E|Y,X_C,X_S)P(Y|X_C)$ & F        \\ \hline
CG7   & $P(X_C)$                   & C        \\
      & $P(X_S)$                   & C        \\
      & $P(X_E|Y,X_S)P(Y|X_C)$     & F       
\end{tabular}
\captionsetup{justification=centering}
\caption{Joint approach: factors and scenarios for CG3, CG5, CG6, CG7}
\label{tab:updatedt3}
\end{table}

\noindent{\textbf{Structural Modelling of features in scenario \textbf{C}, \textbf{D}, \textbf{E}, \textbf{F}}.

For $v_i\in X$, $P(v_i|\parent{v_i})$ may be modelled as SCM $f_{\theta_{v_i}}$, and $f_{\theta_{v_i}}$ is some universal function approximator, such as a feed forward neural network, with parameters $\theta_{v_i}$. Where $f_{\theta_{v_i}}:=\hat{p}(v_i)$, and $p(v_i)$ is the empirical sample of $v_i$, we use the maximum mean discrepancy (MMD) \cite{kernel_two_sample_test_mmd} to match the estimate $\hat{p}(v_i) \stackrel{i.i.d.}{\sim} p(v_i)$. For some feature map $\Phi: \mathcal{X}\rightarrow \mathcal{H}$, where $\mathcal{H}$ is a reproducing kernel Hilbert space (RKHS), MMD computes the mean distance between feature embeddings of distributions $P$ and $Q$:
\begin{equation}
\textbf{MMD}(P\parallel Q)=\| \mathbb{E}_{X\sim P}[\Phi(X)]-\mathbb{E}_{Y\sim Q}[\Phi(Y)] \|_{\mathcal{H}}.
\label{eq:mmd_definition}
\end{equation}

If $\textbf{MMD}(P\parallel Q)=0$, the distributions $P,Q$ are identical. A kernel $k$ is a function that computes the dot product of $X,X'$ in feature space: $k(x,x')=\Phi(x)^T\Phi(x')$. For empirical samples $x\in X,x'\in X'$, we can derive an estimate $\hat{\textbf{MMD}}(X\parallel X')$ by computing $k(x,x')$ for an RKHS kernel such as the radial basis function (RBF) kernel $k_{\sigma_x}$. For bandwidth $\sigma_x\in \mathbb{R}^{+}$,

\begin{equation}
k_{\sigma_x}(x,x')=\exp\Bigg[\frac{-\|x-x'\|^{2}}{2\sigma_x}\Bigg].
\label{eq:rbf_kernel}
\end{equation}

We now move to a detailed explanation of the estimation procedure in scenarios \textbf{A}-\textbf{F}. We employ notations given in Table \ref{tab:mmd_method_notation}.

\begin{table}[H]
\centering
\renewcommand{\arraystretch}{1.5}
\begin{tabular}{| w{r}{2.5cm}| m{5.5cm} |} 
\hline
\textbf{Notation}                           & \textbf{Meaning}                                          \\ \hline
$P(X_A)$ & Distribution of random variable $X_A$ \\ \hline
$P(X_1) \matchdist P(X_2)$        & Distributions of random variables $X_1,X_2$ match \\ \hline
$\textbf{MMD}(P\parallel Q)$        & MMD between distributions $P$, $Q$                \\ \hline
$p(x_a) $        & Empirical samples $x_a$ from distribution $P(X_A)$\\ \hline
$x_a\in D_u $        & Empirical samples $x_a\in X_A$ in unlabelled sample $D_u$ \\ \hline
$\hat{\textbf{BCE}}(p\parallel q)$       & Sample Binary Cross Entropy \cite{deeplearning_book} between realisations from $P, Q$ \\ \hline
$\hat{\textbf{MMD}}(p\parallel q)$ & Sample MMD between realisations from $P, Q$           \\ \hline

\end{tabular}
\captionsetup{justification=centering,width=0.98\textwidth}
\caption{Notations used in estimation procedures, scenario \textbf{A}-\textbf{F}}
\label{tab:mmd_method_notation}
\end{table}

\noindent{\textbf{Scenario A: $P(v_i|\parent{v_i})=P(Y)$}

\noindent The factor $P(Y)$ corresponds to a root node in the DAG, and we consider $P(Y)$ as a random variable, rather than SCM. For $Y\in\{1,2,\dots,K\}$, where $y_i$ counts the number of occurrences of value $y_i$ in $n$ observations, and $\pi_i$ is the probability that $y_i$ occurs in a single observation, $Y$ is modelled using the multinomial distribution with probability mass function (PMF):

\begin{equation}
    P(Y|\pi,n)=n!\prod\limits_{i=1}^{K}\frac{\pi_i^{y_i}}{y_i!}
\end{equation}

\noindent The labelled sample, consisting of $n_l$ observations, may be used to estimate $\hat{\pi_i}=\frac{y_i}{n_l}$ via maximum likelihood. We can then draw samples from a categorical distribution with parameters $\pi=(\pi_1,\pi_2,\dots,\pi_K)$. Intuitively, for two classes this reduces to a Bernoulli random variable.\\

\noindent{\textbf{Scenario B: $P(v_i|\parent{v_i})=P(Y|X_K,\tdots)$}}

\noindent For brevity, we write $X_C=\{X:X\in\parent{Y}\}$, so that $P(Y|X_K,\tdots)=P(Y|X_C)$. Denote $f_{\theta_Y}:=\hat{P}(Y|X_C)$. We fit $Y$ against $X_C$ from the labelled pairs. As such, this is a fully-supervised learning problem. If there are two classes, we use the Binary Cross Entropy loss over the empirical sample, employing the following objective for update of $\theta_Y$:

\begin{align*}
\theta_Y=\underset{\theta_Y}{\argmin}  \underset{(x_c,y)\in \{D_l\}}{\hat{\textbf{BCE}}}&\Big[y\parallel f_{\theta_Y} ({x_c})\Big].
\end{align*}

\noindent{\textbf{Scenario C: $P(v_i|\parent{v_i})=P(X_I)$}

\noindent Denoting our model $f_{\theta_{X_i}} := \hat{P}(X_I)$, our theoretical intuition is that if the MMD distance between the true distribution $P(X_I)$ and our model $\hat{P}(X_I)$ is zero, then $P(X_I)\matchdist \hat{P}(X_I)$. Therefore, our goal is to find the model parameters $\theta_{X_i}$ of $f_{\theta_{X_i}}$ which minimise the sample MMD between the empirical distribution of $X_I$, and samples estimated from $f_{\theta_{X_i}}$. Following the SCM framework, $f_{\theta_{X_i}}$ is a function of the independent noise term $N_{X_i}$ only. Since the causal mechanism has no dependence on a latent $Y$, all samples are observed in labelled and unlabelled data. This provides the following objective:

\begin{align*}
\theta_{X_i}=\underset{\theta_{X_i}}{\argmin}  \underset{x_i\in \{D_l,D_u\}}{\hat{\textbf{MMD}}}&\Big[p(x_i)\parallel f_{\theta_{X_i}} (N_{X_i})\Big].
\end{align*}


\noindent{\textbf{Scenario D: $P(v_i|\parent{v_i})=P(X_I|X_K,\tdots)$}

\noindent Using shorthand notation so that $X_C=\{X:X\in\parent{X_I}\}$, $P(X_I|X_K,\tdots)=P(X_I|X_C)$. Similar to Scenario C, where the feature $X_I$ is a root node, we do not have to deal with missing labels, $f_{\theta_{X_i}} := \hat{P}(X_I|X_C)$ can be estimated from paired observations $(x_i,x_c)$ in unlabelled and labelled data. $f_{\theta_{X_i}}$ is a function of the noise term $N_{X_i}$ and parent features $X_C\in\parent{X_i}$. Since a perfect model of the causal mechanism $f_{\theta_{X_i}}:=\hat{P}(X_I|X_C)$ equates to the following,

\begin{align*}
\textbf{MMD}&\Big[P(X_I|X_C)P(X_C)\parallel \hat{P}(X_I|X_C)P(X_C)\Big] = 0,
\end{align*}

\noindent our goal is to find model parameters $\theta_{X_i}$ which minimise the sample MMD between the empirical joint distribution $P(X_i,X_c)$ and the causal factorisation $\hat{P}(X_I|X_C)P(X_C)$:

\begin{align*}
\theta_{X_i}=\underset{\theta_{X_i}}{\argmin}  \underset{(x_i,x_c)\in \{D_l,D_u\}}{\hat{\textbf{MMD}}}&\Big[p(x_i,x_c)\parallel f_{\theta_{X_I}} (N_{X_I},x_c)p(x_c)\Big].
\end{align*}

\noindent{\textbf{Scenario E: $P(v_i|\parent{v_i})=P(X_I|Y,\tdots)$}

\noindent Since $f_{\theta_{X_i}}:=\hat{P}(X_I|Y,\tdots)$ is a function of root node $Y$, we seek to benefit from unlabelled and labelled samples. First, we discuss the case where there are no features $X_S\in\parent{X_i}$, so that $P(v_i|\parent{v_i})=P(X_I|Y)$. The optimisation procedure iterates over minibatches from labelled and unlabelled samples separately. For labelled sample, we use $f_{\theta_{X_i}}$ to estimate $P(X_I|Y)$. If the estimate $\hat{P}(X_I|Y)$ is optimal, then $\textbf{MMD}\Big[P(X_I|Y)P(Y)\parallel \hat{P}(X_I|Y)P(Y)\Big]=0$, which suggests the following training objective for labelled data:

\begin{align*}
\theta_{X_i}=\underset{\theta_{X_i}}{\argmin} \underset{(y,x_i)\in \{D_l\}}{\hat{\textbf{MMD}}}&\Big[p(x_i,y)\parallel f_{\theta_{X_i}}(y,N_{X_i})p(y)\Big].
\end{align*}

\noindent As $f_{\theta_{X_i}}$ must be provided some $Y$ during training, we must modify this for the unlabelled batch: unlabelled sample is used to estimate the marginal $P(X_I)$ from $f_{\theta_{X_i}}$, rather than $P(X_I|Y)$.  We create a sample $(X_I,Y^*)$, where $X_I\in D_u$, and $Y^*$ is a bootstrapped sample randomly drawn from labels $Y\in D_l$. As $Y^*$ are randomly drawn with replacement, this conforms to the probabilistic measure which renders $Y^*$ and $X_I$ independent. Comparing the bootstrapped and original sample, we expect the following to hold:

\begin{equation*}
\sum_{Y^*\in D_l} P(X_I|Y^*)P(Y^*) \matchdist \sum_{Y\in D_l} P(X_I|Y)P(Y).
\end{equation*}

\noindent If $\hat{P}(X_I)$ is optimal, we have $\textbf{MMD}\Big[P(X_I)\parallel \hat{P}(X_I)\Big]=0, X_I\in D_u
$.
\noindent Hence, we use the following objective function to estimate the marginal $P(X_I)$ from unlabelled data:

\begin{align*}
\theta_{X_i}=\underset{\theta_{X_i}}{\argmin}  \underset{x_i\in D_u,y^*\in D_l}{\hat{\textbf{MMD}}}&\Big[p(x_i)\parallel f_{\theta_{X_i}}(N_{X_i},y^*)p(y^*)\Big].
\end{align*}
If we have some $X_S\in\parent{X_I}$, then we need to pair each observation with an appropriate sample from $X_S$. We modify the the labelled objective as follows:
\begin{align*}
\hspace*{-1cm}
\theta_{X_i}=&\underset{\theta_{X_i}}{\argmin} \underset{(y,x_i,x_s)\in \{D_l\}}{\hat{\textbf{MMD}}}\\&\Big[p(x_i,y,x_s)\parallel f_{\theta_{X_i}}(N_{X_i},y,x_s)p(y)p(x_s)\Big].
\end{align*}
And we modify the unlabelled objective as follows:
\begin{align*}
\hspace*{-1cm}
\theta_{X_i}&=\underset{\theta_{X_i}}{\argmin}  \underset{\substack{(x_i,x_s)\in D_u,y^*\in D_l}}{\hat{\textbf{MMD}}}\\&\Big[p(x_i,x_s)\parallel f_{\theta_{X_i}}(N_{X_i},y^*,x_s)p(y^*)p(x_s)\Big].
\end{align*}

\noindent{\textbf{Scenario F: $P(X_I|Y,\tdots)P(Y|\tdots)$}

\noindent We illustrate this method on the causal structure given in Figure \ref{fig:mb_topologies}, for CG3 $X_C\rightarrow Y \rightarrow X_I$. For more complicated structures, ie if $X_I$ is a function of any $X_S,X_C$, we modify the SCM for $X_I$, $f_{\theta_{X_i}}$, and use appropriate samples from unlabelled and labelled data. The key contribution of this section is to show that if $Y$ is not a root node, and hence the DAG contains some $X_C\rightarrow Y$ as well as $Y\rightarrow X_I$, it may be advantageous to estimate structural model parameters for $X_I,Y$ together, because $Y$ mediates the relationship between $X_C$ and $X_I$: this is the motivation for employing Scenario \textbf{F} in the joint approach, instead of using scenario \textbf{B} and \textbf{E} in the disjoint approach. Beginning with the causal factorisation $P(X_C,Y,X_I)=P(X_I|Y)P(Y|X_I)P(X_C)$, denote structural models $g_{\theta_Y}(X_C):=\hat{P}(Y|X_C)$, and $f_{\theta_{X_i}}(N_{X_i},Y):=\hat{P}(X_I|Y)$. Under the disjoint approach, $g_{\theta_Y}$ is modelled as a regression/classification function via scenario \textbf{B}, using only labelled samples $\{X_C,Y\}\in D_l$, and $f_{\theta_{X_i}}$ is modelled as scenario \textbf{E}. Combining these structural models, we derive an expression for $\hat{P}(X_I|X_C)$:

\begin{align*}
\hat{P}(X_I|X_C)=\int_Y\hat{P}(X_I|Y)\hat{P}(Y|X_C).
\end{align*}

However, given DAG $X_C\rightarrow Y\rightarrow X_i$, $\hat{P}(X_I|X_C)$ should depend on $Y$, and this dependence is not well captured if we optimise $g_{\theta_Y}$ and $f_{\theta_{X_i}}$ separately. In this instance, we expect the composite causal mechanism $f \circ g$ to be modelled for data in the labelled dataset only. In scenario \textbf{F}, we wish to capture information about $Y$ from the unlabelled $(x_i,x_c)$ pairs. In this case, we use the Gumbel-Softmax \cite{jang2017categorical} trick to train models for $f,g$ together. In our disjoint method, we used bootstrapped $y^{*}$ to match unlabelled $x_i\in D_u$. In contrast, we now instead sample from the estimate $\hat{P}(y|x_c), x_c\in D_u$ to exploit information in unlabelled $(x_i,x_c)$. Using Gumbel-Softmax, hard-labelled estimates $\hat{y}\sim\hat{P}(Y|X_C)$ may be drawn by sampling $G^i\sim-\log(-\log(\text{Uniform}(0,1)))$, adding to $\pi_y$, which are the normalised outputs from $g_{\theta_Y}:=\hat{P}(Y|X_C)$, and finally taking argmax: $\hat{y}=\argmax\{\theta_y + G^{i}\}$. By using this estimate, we link the causal mechanism $f \circ g$ for paired unlabelled observations $(X_C,X_I)\in D_u$:

\begin{align*}
P(X_I,X_C)  =& \int\limits_Y P(X_I|Y)P(Y|X_C)P(X_C) dY, \\
=& P(X_I|X_C)P(X_C), \{X_I,X_C\}\in D_u.
\end{align*}

Considering that $\textbf{MMD}(p\parallel q)=0$ iff $p \matchdist q$, and according to the causal factorisation each factor shares no information, then the disjoint terms must match, and we expect to improve $\hat{P}(Y|X_C): \textbf{MMD}\Big[P(X_I|Y)P(Y|X_C)||\hat{P}(X_I|Y)\hat{P}(Y|X_C)\Big]=0.$ However, this is a special case which requires the existence of some $X_C$ in addition to $X_E$. If there is no $X_C$, such as in CG4 and CG2, only the disjoint method could be used to produce a generative model of $P(X,Y)$.

\textbf{Sampling data $D_G$ from our model $G:=\hat{P}(X,Y)$}. Once we have optimised the structural models of $G$, we perform ancestor sampling to generate novel data $D_G=(x',y')$ from $G$. This procedure is straightforward: we generate samples of root node variables from structural models $f_{\theta_v}(N_{v})$, and these samples are then used as input to estimate all subsequent variables. We refer to the generated data as $D_G$. The number of examples drawn from $G$ is the same as the number of unlabelled data, $D_u$, as recorded in Table \ref{tab:synthetic_partition_splits} and Table \ref{tab:real_partition_splits}.

\textbf{Training the classifier}. For the classifier $\mathcal{C}:\mathcal{X}\rightarrow \mathcal{Y}$, we first train on labelled pairs $(x,y)\in D_l$. We can think of this as `pre-training'. Then, we augment the sample with $D_G$ and use $D'=D_G\cup D_l$ to perform further training.

\section{Experiments}

To develop benchmarks for the improvement from an SSL method, we train a classifier on labelled data only. We call this model the partial supervised classifier, or P-SUP. We train a separate classifier using a modified dataset where all of the labels from the unlabelled data are given to the model. We call this model the fully supervised classifier, or F-SUP. We expect F-SUP to indicate an upper bound on the performance achievable by any SSL method. For each dataset, we take $n=100$ examples, and report the average difference in classification accuracy relative to the P-SUP model.

\textbf{Neural Network Implementation Details}. We employ two neural network architectures for our experiments, both using rectified linear unit (ReLU) activation. The first architecture is a 3-layer multilayer perceptron (MLP) with hidden layer size 100. This architecture is used in all benchmark models, and in the classifier network for our method. The second architecture is identical except the hidden layer size is 50. This architecture is used for all SCM models.

We experimented with a variety of architectures which varied over hidden layer size 50, 100 or 200, and either 1 hidden layer or 3 hidden layers. We found that a simple MLP with either 100 or 50 neurons was easiest to train and exhibited most consistent performance across models.

During training, we implement early stopping regularisation if classification accuracy over $D_v$, the validation partition, does not increase over 10 epochs. Recall that the MMD loss requires a kernel satisfying the RKHS property, in equation \ref{eq:mmd_definition}. K is a mixture of five RBF kernels, as given in Equation \ref{eq:rbf_kernel}. To derive all kernels, set $\sigma_x$ to the median pairwise distance between all points $x\in X_i$. Then the mixture, with each component indexed by $n=[1,\dots,5]$, and $\sigma_{x_n}=2^{n-3}\sigma_x$, is given by:

\begin{equation*}
    K(x,x')=\sum\limits_{n=1}^{5} k_{\sigma_{x_n}}(x,x').
\end{equation*}

\textbf{Our causal SSL method}. We demonstrate two implementations of our method, corresponding to the joint / disjoint modelling approaches. The joint method, which uses the the Gumbel-Softmax to jointly optimise modules under identification of scenario \textbf{F}, is denoted Gumbel-conditional semi-supervised GAN (GCGAN-SSL). The disjoint method, which identifies scenarios \textbf{B}, \textbf{E} instead of \textbf{F}, is denoted conditional semi-supervised GAN (CGAN-SSL).

\textbf{Benchmark SSL models for comparison}. We report the performance of benchmark SSL models as tabulated in Table \ref{tab:ulab_acc_synthetic}: SSL-GAN \cite{ssl_gan_original}, Triple-GAN \cite{triplegan_orig}, SSL-VAE \cite{SVAE}, VAT \cite{VAT_orig}, ENT-MIN \cite{entmin_orig}, label propagation/pseudolabeling (L-PROP) \cite{pseudolab_orig}, adaptive semi-supervised feature selection (ADapt-SSFS) \cite{adapt_ssfs_2018}, and semi-supervised feature analysis (SSFA-Cor) \cite{ssfa_cor_2016}. In SSL-VAE, we use a latent embedding of dimension 5. Data analysis was undertaken using The University of Melbourne’s Research Computing Services, supported by the Petascale Campus Initiative.

\subsection{Synthetic data}

We first demonstrate the utility of our approach on seven different synthetic datasets, each of which corresponds to the causal graphs CG1-CG7, as depicted in Figure \ref{fig:markov_blanket_full} and Figure \ref{fig:mb_topologies}. Each dataset conforms to a different Markov Blanket over $Y$, allowing us to test the ideas proposed in the current work.

\textbf{How we generate synthetic data}. For each DAG, our goal is to generate synthetic data with a nonlinear decision boundary to be used for a semi-supervised binary classification task. In the basic method, each dataset instance consists of $n=2080$ cases, which are split into labelled, unlabelled, validation and test partitions. This number is changed if we want to include more unlabelled cases, as per Table \ref{tab:synthetic_partition_splits}. The generative processes for CG1-CG7 are explained sequentially, starting with CG1. 

\begin{enumerate}
    \item{\textit{CG1: }For dataset instance $i$, $X_C \sim N(0,\Sigma_{s_i})$, $\Sigma_{s_i} = \begin{bmatrix} s_i & 0  \\ 0 & s_i^{-1}  \end{bmatrix}$, and each $s_i$ is a single sample drawn from $\text{Uniform}(1,2)$, ie $1\leq s_i \leq 2$. We use quadratic feature map $\Phi_i(X_C):\mathbb{R}^2\rightarrow \mathbb{R}$ to set a decision boundary $P(Y|X)$, where the elements of $\Phi_i(X_C)$ are randomly assigned for each $i$. Write $X_C=\begin{bmatrix} X_{C_1} \\ X_{C_2} \\ \end{bmatrix}$. $\Phi_i$ is determined by 6 scalars $a,b,c,d,e,f$: $
    \Phi_i(X_C)=a_iX_{C_1}^2+b_iX_{C_2}^2+c_iX_{C_1}+d_iX_{C_2}+e_iX_{C_1}X_{C_2} + f_i$.  Set $f_i=0, \forall i$, and each $a_i,b_i,c_i,d_i,e_i$ is a separate single observation from distribution $u\sim \text{Uniform}(0,1)$. Now, we use the feature map $\Phi_i(X_C)$ to parameterise a Bernoulli distribution, assigning the label $Y$ from $X_C$. $\sigma(x_c,\mu_{\Phi_i})$ is a Sigmoid function centered at $\mu_{\Phi_i}$: $\sigma(x_c,\mu_{\Phi_i})=\frac{1}{1+e^{-(\Phi_i{(x_c)}-\mu_{\Phi_i})}}$. $\mu_{\Phi_i}$ is the sample mean embedding, ie  $\mu_{\Phi_i}=\frac{1}{2080}\sum\limits_{x_c\in X_C}\Phi_i(x_c)$. Then write $p_{\Phi_i}(x_c)=\sigma(x_c,\mu_{\Phi_i})$, and each $y\in Y$ is a single sample from Bernoulli($p_{\Phi_i}(x_c)$), corresponding to each $x_c\in X_C$. For our synthetic data, we ideally want to keep $P(Y=0)\approx P(Y=1)\approx 0.5$. For each random parameterisation, we synthesise 2080 cases and keep the dataset if $0.45\leq P(Y) \leq 0.55$. If this is not the case, we discard the dataset and randomly draw new parameters. This procedure is repeated until we have n=100 instances for CG1. }
    \item{\textit{CG2: }Set $Y\sim\text{Bernoulli}(0.5)$, and for each dataset instance $i$, $w_i$ is a single sample drawn from $\text{Uniform}(4,6)$, and $\Sigma_{w_i}$ is defined analogously to $\Sigma_{s_i}$ for CG1. The causal mechanism $f_{X_E}$ is a nonlinear transform of $N_{X_E} \sim N([0,0],\mathbb{I}_2)$, resulting in a nonlinear decision boundary between classes \cite{pclass_2001}. Write $X_E=\begin{bmatrix} X_{E_1} \\ X_{E_2} \\ \end{bmatrix}$. Define a template causal mechanism $f_{X_E}^{\textbf{T}}(N_{X_E},Y,A)$, to be used in CG2-CG7, in Equation \ref{eq:template_cmech}, where $A\in\mathbb{R}$ is an offset in the second dimension:\\
    \begin{adjustbox}{minipage=\linewidth,scale=0.8,right,lap=-0.2\width}
    \begin{minipage}{1.2\linewidth}
    \begin{equation}
    \medmuskip=2mu   
    \thickmuskip=3mu 
    \renewcommand\arraystretch{1.5}
    f_{X_E}^{\textbf{T}}(N_{X_E},Y,A) =
    \left\{\begin{array}{@{}l@{}}
    \Sigma_{w_i} N_{X_E} + \begin{bmatrix} 0\\4\cos\frac{N_{X_{E_2}}}{2}\end{bmatrix}, \;\;\;\;\;\;\;\;Y=0\\\\\
    \Sigma_{w_i} N_{X_E} + \begin{bmatrix} 0\\ 4\cos\frac{N_{X_{E_2}}}{2}+A\end{bmatrix},\;\;Y=1
    \end{array}\right.
    \label{eq:template_cmech}
    \end{equation}
    \end{minipage}
    \end{adjustbox}\\
    
    Write $f_{X_E}^{\textbf{CG}X}$ as the causal mechanism of $f_{X_E}$ for dataset $\textbf{CG}X$. Set $f_{X_E}^{\textbf{CG2}}(N_{X_E},Y):=f_{X_E}^{\textbf{T}}(N_{X_E},Y,\frac{w_i}{2})$. The decision boundary is visibly curved, as depicted in Figure \ref{fig:cg2_example_curve}.}
    \item{\textit{CG3: }$X_C$ and $Y$ are generated as per CG1, and $X_E$ is generated as per CG2.}
    \item{\textit{CG4: }For dataset instance $i$, $X_S \sim N([0,0],\Sigma_{t_i})$, each $t_i$ is a single sample drawn from $\text{Uniform}(1,2)$, with $\Sigma_{t_i}$ generated analogously as for CG1. Set $Y\sim\text{Bernoulli}(0.5)$. We set the causal mechanism $f_{X_E}^{\textbf{CG4}}(N_{X_E},Y,X_S):=f_{X_E}^{\textbf{T}}(N_{X_E},Y,\frac{t_i}{2}+\frac{w_i}{2})+X_S$.}
    \item{\textit{CG5: }$X_C$ and $Y$ are generated as per CG1. For $X_E$, define the causal mechanism $f_{X_E}^{\textbf{CG5}}(N_{X_E},Y,X_C):=f_{X_E}^{\textbf{T}}(N_{X_E},Y,\frac{s_i}{2}+\frac{w_i}{2})+X_C$.}
    \item{\textit{CG6: }$X_C,Y$ are generated as per CG1, and $X_S$ is generated as per CG4, and set $f_{X_E}^{\textbf{CG6}}(N_{X_E},Y,X_C,X_S):=f_{X_E}^{\textbf{T}}(N_{X_E},Y,\frac{s_i}{2}+\frac{t_i}{2}+\frac{w_i}{2})+X_C+X_S$.}
    \item{\textit{CG7: }$X_C,Y$ are generated as per CG1, and $X_S$ is generated as per CG4. For $X_E$, set $f_{X_E}^{\textbf{CG7}}(N_{X_E},Y,X_S):=f_{X_E}^{\textbf{T}}(N_{X_E},Y,\frac{t_i}{2}+\frac{w_i}{2})$.}
\end{enumerate}


$\textbf{Partition splits for synthetic data}$. All experiments on synthetic data use n=40 labelled for training, and n=40 cases for validation. Keeping these amounts constant, we conducted a number of experiments with varying amounts of unlabelled data  $|D_u|=\{1000,5000,10000\}$ randomly split into partitions given in Table \ref{tab:synthetic_partition_splits}. The test partition $D_t$ is used to evaluate accuracy once at end of training. In our experiments, we observe similar performance between $D_u$ and $D_t$, but we only report accuracy over $|D_u|=10000$ for brevity. $D_G$ denotes the number of cases that we create via our causal generative models CGAN-SSL / GCGAN-SSL.

\begin{table}[H]
\centering
\begin{tabular}{|c|c|c|c|c|c|}
\hline
\textbf{Partition} & Labelled & Unlabelled & Validation & Test & Generated \\ \hline
\textbf{Notation} & $D_l$ & $D_u$ & $D_v$ & $D_t$ & $D_G$ \\ \hline
\textbf{Size} & 40 & 1000 & 40 & 1000 & 1000 \\ \hline
\textbf{Size} & 40 & 5000 & 40 & 5000 & 5000 \\ \hline
\textbf{Size} & 40 & 10000 & 40 & 10000 & 10000 \\ \hline
\end{tabular}
\captionsetup{justification=centering,width=0.5\textwidth}
\caption{Partition splits used for CG1-CG7}
\label{tab:synthetic_partition_splits}
\end{table}

\begin{table}[H]
\centering
\begin{tabular}{|r|c|c|c|c|c|}
\hline
Source Data & $D_l$ & $D_u$ & $D_v$ & $D_G$ \\ \hline
Breast Cancer Wisconsin & 10 & 424 & 10 & 424 \\ \hline
Sachs & 10 & 7446 & 10 & 7446 \\ \hline
\end{tabular}
\captionsetup{justification=centering,width=0.8\linewidth,font=small}
\caption{Partition splits for real data experiments: see Table \ref{tab:synthetic_partition_splits} for partition notation}
\label{tab:real_partition_splits}
\end{table}

\begin{figure}
\centering
\begin{adjustbox}{minipage=\linewidth,scale=0.6,center,lap=-0.05\width}
\centering
\includegraphics[trim = 120 53 65 330, clip, width=\textwidth,scale=0.05]{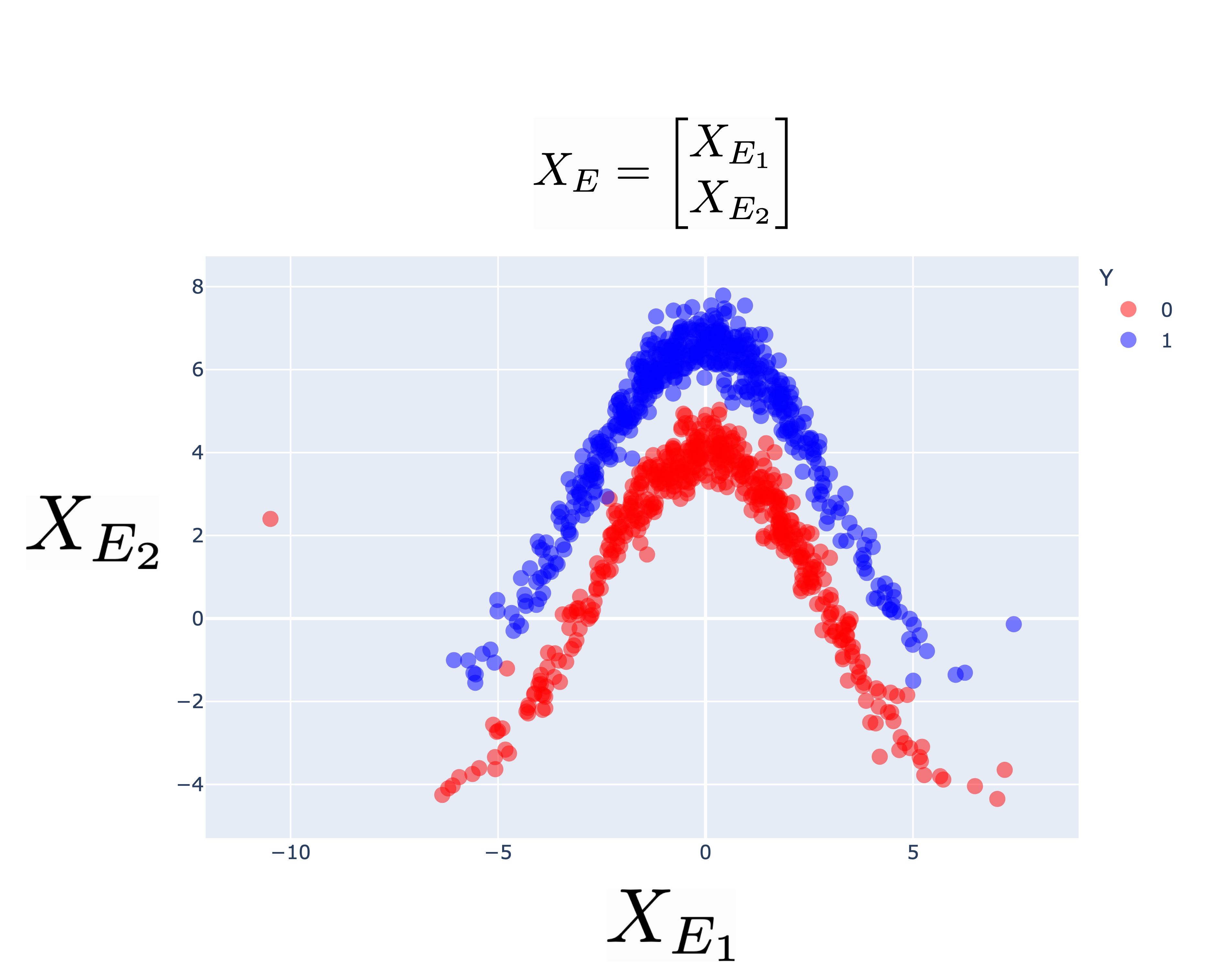}
\end{adjustbox}
\captionsetup{justification=centering,width=\textwidth}
\caption{Curved decision boundary between classes for CG2}
\label{fig:cg2_example_curve}
\end{figure}

\textbf{SSL performance on synthetic data}. We tabulate classification accuracy over $D_u$ in Table \ref{tab:ulab_acc_synthetic}, with corresponding box plots in Figure \ref{fig:bplot_synthetic_performance}. CG1 illustrates the anticausal SSL conjecture in benchmark methods \cite{anticausal_learning}. In contrast, our method demonstrates some improvement over baseline, although the variability is also higher, as depicted in Figure \ref{fig:bplot_synthetic_performance} (a). The results for CG2 suggest that in general this decision boundary is difficult to learn. By keeping a similar causal mechanism $f_{X_E}$ across CG2-CG7, we illustrate the potential of extra features to improve the model. CG3, CG5 and CG7 illuminate the strongest support for causality in SSL. By parameterising a composite causal mechanism over $Y$ for unlabelled data, the joint Gumbel-Softmax method GCGAN-SSL seems able to exploit information in unlabelled data more effectively than the disjoint method CGAN-SSL. CG5 illustrates that a disjoint causal approach via CGAN-SSL may actually worsen model performance in this instance. To a lesser extent, this notion is also reflected in CG6: GCGAN-SSL performs better than CGAN-SSL. CG4 shows a weaker but still compelling case for a causal approach to SSL: although SSL-GAN is superior, CGAN-SSL is the next-best performing model.

\begin{sidewaysfigure}
\begin{adjustbox}{minipage=\textwidth,scale=0.8}
\begin{table}[H]
\setlength{\tabcolsep}{0.3em}
\renewcommand{\arraystretch}{0.75}
\centering

\begin{tabular}{rccccccccccc}
\toprule
{} &                                                                   KEY &                     CG1 &                     CG2 &                      CG3 &                     CG4 &                     CG5 &                      CG6 &                      CG7 &                 BCANCER &                     RAF \\
\midrule
F-SUP      &        \textcolor{FULLY_SUPERVISED_CLASSIFIER}{\LARGE $\blacksquare$} &           4.762 ± 3.970 &           8.989 ± 4.869 &           22.200 ± 7.398 &          18.281 ± 6.956 &          18.296 ± 6.407 &           22.945 ± 4.763 &           27.724 ± 4.986 &           4.495 ± 3.108 &           6.429 ± 5.538 \\
CGAN-SSL   &   \textcolor{CGAN_BASIC_SUPERVISED_CLASSIFIER}{\LARGE $\blacksquare$} &  \textbf{0.218 ± 3.228} &           2.787 ± 5.167 &           13.637 ± 7.542 &           3.023 ± 7.107 &           2.211 ± 6.751 &            7.475 ± 5.334 &           10.483 ± 5.980 &  \textbf{1.898 ± 3.100} &           1.969 ± 6.581 \\
GCGAN-SSL  &  \textcolor{CGAN_GUMBEL_SUPERVISED_CLASSIFIER}{\LARGE $\blacksquare$} &                       - &                       - &  \textbf{16.176 ± 8.923} &                       - &  \textbf{9.507 ± 6.877} &  \textbf{14.059 ± 7.229} &  \textbf{22.073 ± 6.799} &           1.699 ± 3.043 &           0.274 ± 6.240 \\
SSL-GAN    &                            \textcolor{SSL_GAN}{\LARGE $\blacksquare$} &           0.060 ± 2.876 &           0.180 ± 4.151 &            0.123 ± 5.943 &           0.289 ± 6.050 &          -0.722 ± 5.159 &            0.645 ± 4.613 &            0.909 ± 4.427 &           0.154 ± 2.925 &          -1.995 ± 9.176 \\
TRIPLE-GAN &                         \textcolor{TRIPLE_GAN}{\LARGE $\blacksquare$} &          -0.092 ± 3.407 &          -1.098 ± 5.322 &           -5.811 ± 7.554 &          -6.972 ± 7.887 &          -4.430 ± 7.303 &           -1.864 ± 4.974 &           -0.894 ± 4.911 &          -5.433 ± 8.600 &          -2.412 ± 6.549 \\
SSL-VAE    &                            \textcolor{SSL_VAE}{\LARGE $\blacksquare$} &           0.212 ± 3.012 &           0.255 ± 4.416 &           -1.043 ± 6.907 &          -1.151 ± 6.264 &          -1.610 ± 5.506 &           -0.174 ± 4.789 &           -0.404 ± 4.884 &           0.289 ± 2.678 &          -0.279 ± 5.543 \\
VAT        &                                \textcolor{VAT}{\LARGE $\blacksquare$} &          -0.332 ± 3.240 &  \textbf{3.294 ± 4.516} &            6.564 ± 6.958 &  \textbf{5.986 ± 6.596} &           5.703 ± 5.449 &            6.573 ± 5.362 &            8.272 ± 5.754 &           0.100 ± 1.364 &          -0.832 ± 6.521 \\
ENT-MIN    &               \textcolor{ENTROPY_MINIMISATION}{\LARGE $\blacksquare$} &           0.074 ± 2.823 &           0.179 ± 3.234 &           -0.493 ± 4.855 &          -0.100 ± 4.567 &          -1.915 ± 5.192 &            0.202 ± 4.081 &            0.056 ± 3.961 &           0.050 ± 0.979 &           1.368 ± 4.603 \\
L-PROP     &                  \textcolor{LABEL_PROPAGATION}{\LARGE $\blacksquare$} &          -0.017 ± 3.030 &           0.551 ± 6.359 &           -2.480 ± 6.874 &           1.728 ± 6.816 &           0.206 ± 6.763 &            1.328 ± 4.391 &            1.055 ± 4.412 &          -1.694 ± 3.206 &        -10.147 ± 10.727 \\
Adapt-SSFS &                           \textcolor{ASSFSCMR}{\LARGE $\blacksquare$} &                       - &                       - &           3.149 ± 15.598 &        -15.380 ± 15.585 &         -8.357 ± 13.342 &          -8.641 ± 20.326 &          -3.220 ± 15.020 &         -28.786 ± 8.828 &         -12.801 ± 9.885 \\
SSFA-Cor   &                           \textcolor{SFAMCAMT}{\LARGE $\blacksquare$} &          -0.827 ± 4.212 &         -19.619 ± 5.257 &            1.806 ± 7.205 &          -3.635 ± 6.936 &           0.888 ± 6.309 &            7.717 ± 4.731 &            9.027 ± 4.944 &          -4.403 ± 4.335 &  \textbf{2.045 ± 6.120} \\
P-SUP      &      \textcolor{PARTIAL_SUPERVISED_CLASSIFIER}{\LARGE $\blacksquare$} &          62.662 ± 4.184 &          88.894 ± 4.976 &           74.896 ± 7.075 &          79.716 ± 6.911 &          79.801 ± 6.211 &           74.904 ± 4.631 &           70.284 ± 4.434 &          91.915 ± 2.907 &          74.229 ± 5.469 \\
n          &                                                                       &                     100 &                     100 &                      100 &                     100 &                     100 &                      100 &                      100 &                     100 &                     100 \\
\bottomrule
\end{tabular}

\captionsetup{justification=centering,singlelinecheck=false}

\caption{Prediction accuracy for synthetic data ($|D_u|=10000)$) and real data over unlabelled partition, optimal performance over each dataset in $\textbf{bold}$. }
\label{tab:ulab_acc_synthetic}
\end{table}
\end{adjustbox}
\end{sidewaysfigure}

\begin{figure*}
    \begin{adjustbox}{minipage=\linewidth,scale=1.32,center}
        \centering
        \begin{subfigure}[b]{0.32\textwidth}
            \includegraphics[trim = 60 400 400 500, clip, width=\textwidth]{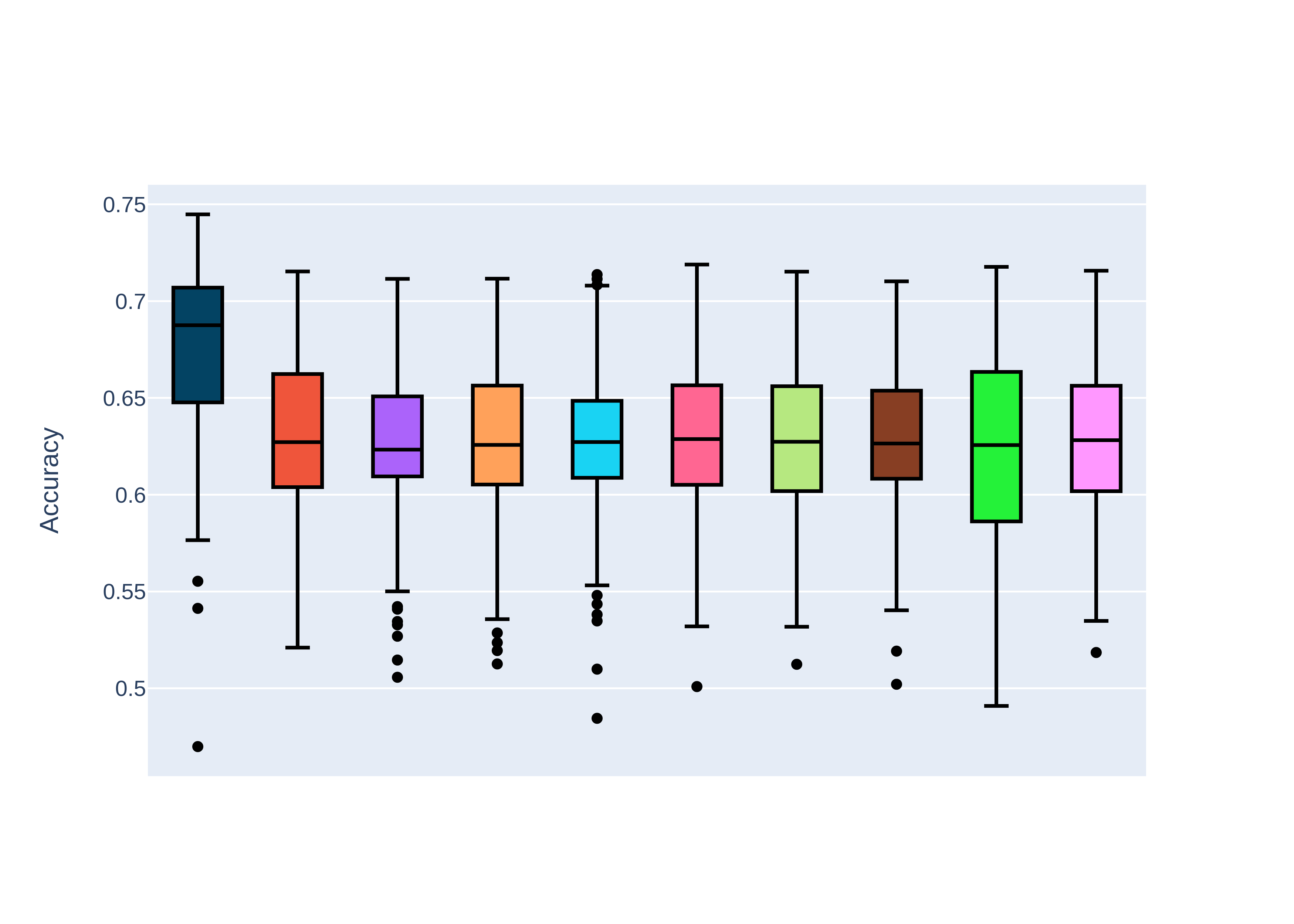}
            \captionsetup{justification=centering,font={footnotesize}}
            \caption{CG1}
        \end{subfigure}
        \begin{subfigure}[b]{0.32\textwidth}
            \includegraphics[trim = 60 400 400 500, clip, width=\textwidth]{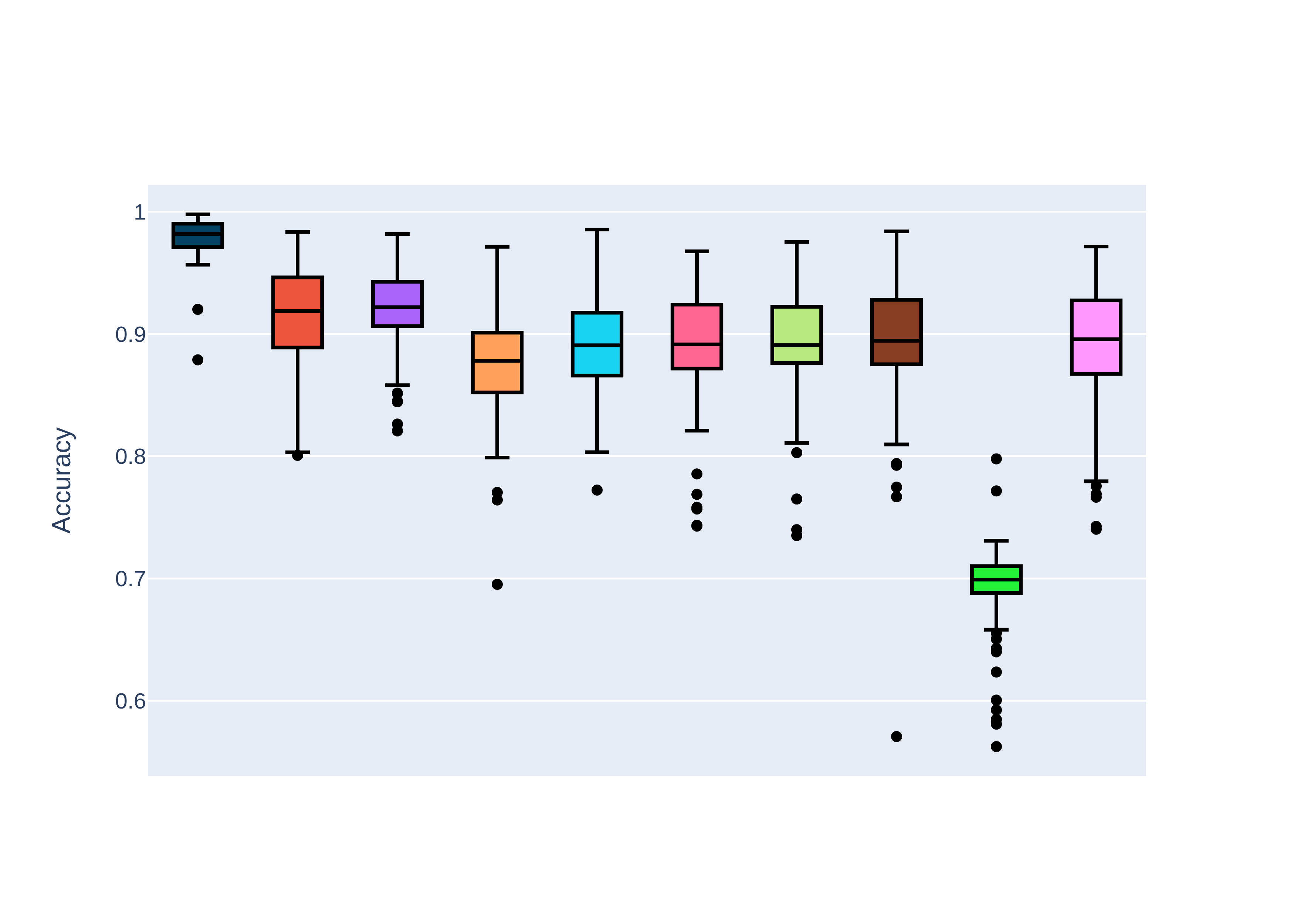}
            \captionsetup{justification=centering,font={footnotesize}}
            \caption{CG2}
        \end{subfigure}
        \begin{subfigure}[b]{0.32\textwidth}
            \includegraphics[trim = 60 400 400 500, clip, width=\textwidth]{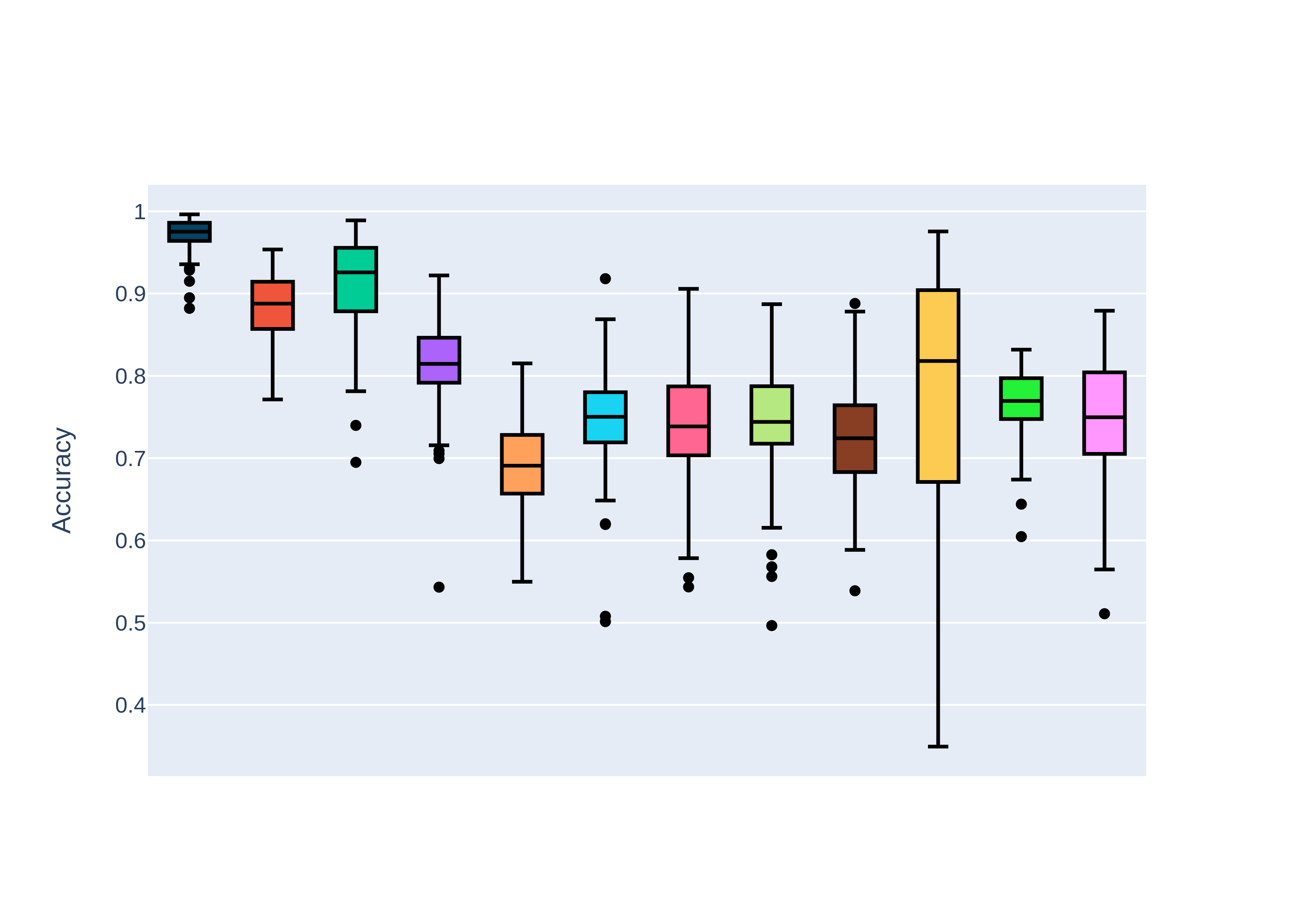}
            \captionsetup{justification=centering,font={footnotesize}}
            \caption{CG3}
        \end{subfigure}\\
        \begin{subfigure}[b]{0.32\textwidth}
            \includegraphics[trim = 60 400 400 500, clip, width=\textwidth]{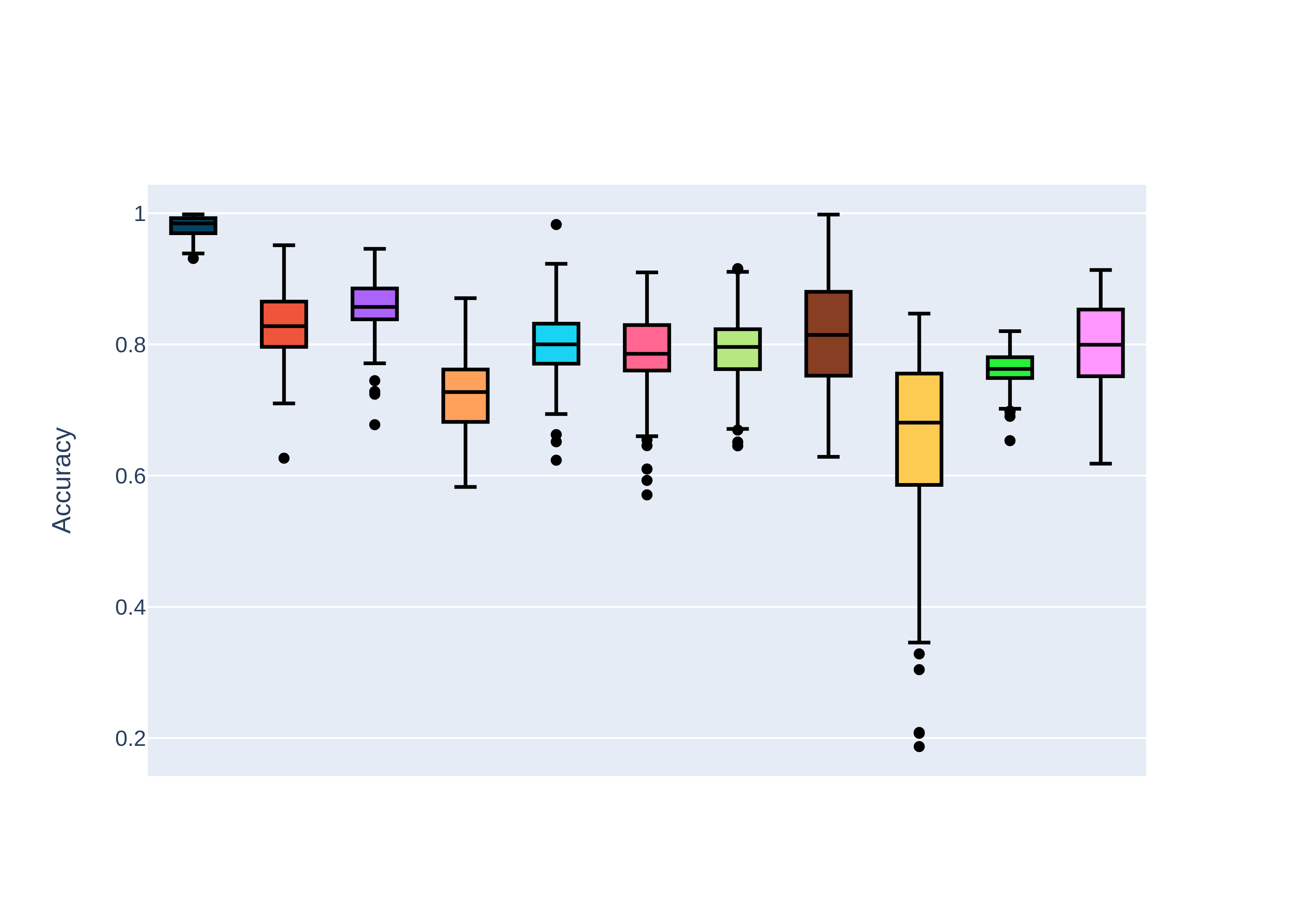}
            \captionsetup{justification=centering,font={footnotesize}}
            \caption{CG4}
        \end{subfigure}
        \begin{subfigure}[b]{0.32\textwidth}
            \includegraphics[trim = 60 400 400 500, clip, width=\textwidth]{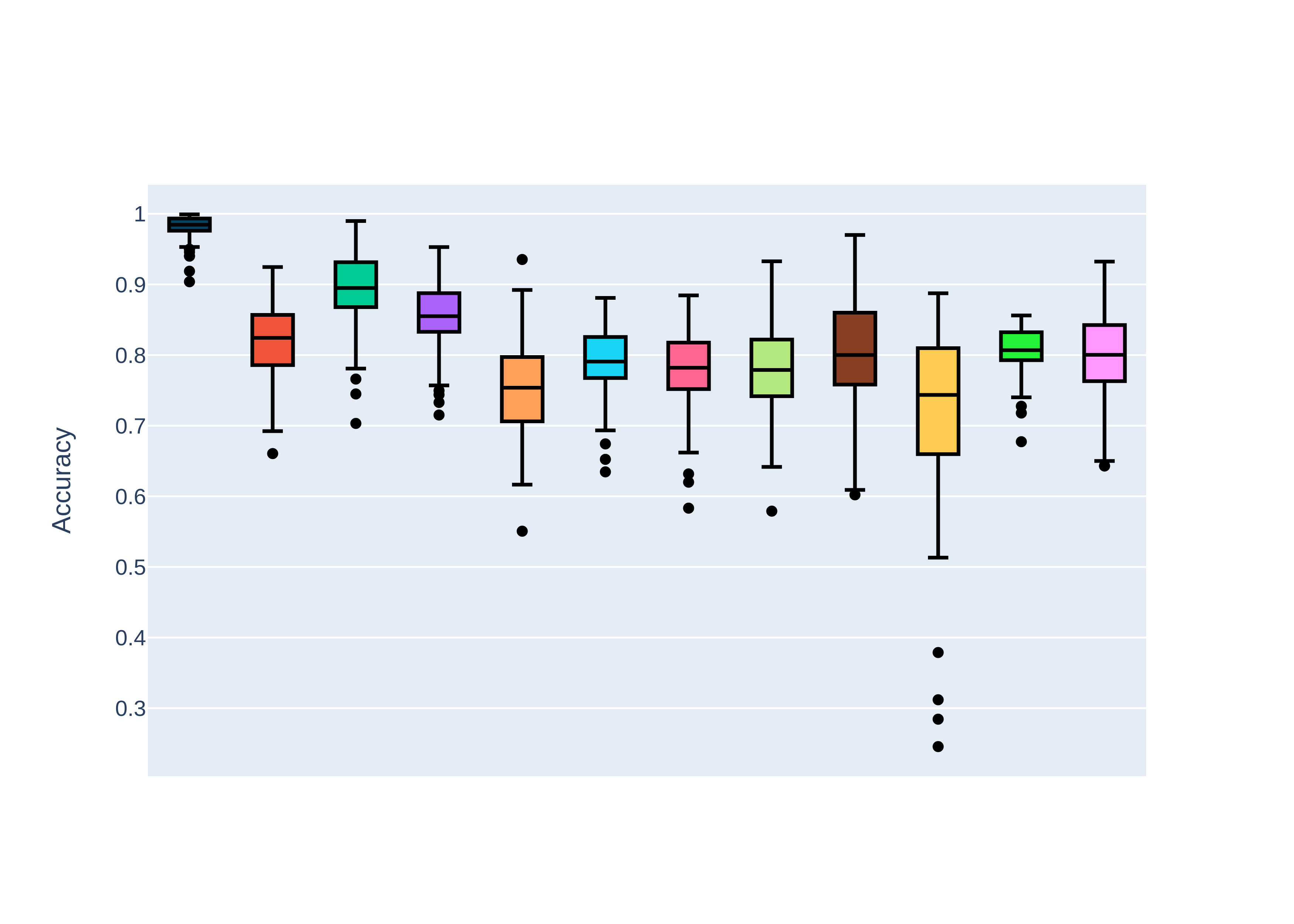}
            \captionsetup{justification=centering,font={footnotesize}}
            \caption{CG5}
        \end{subfigure}
        \begin{subfigure}[b]{0.32\textwidth}
            \includegraphics[trim = 60 400 400 500, clip, width=\textwidth]{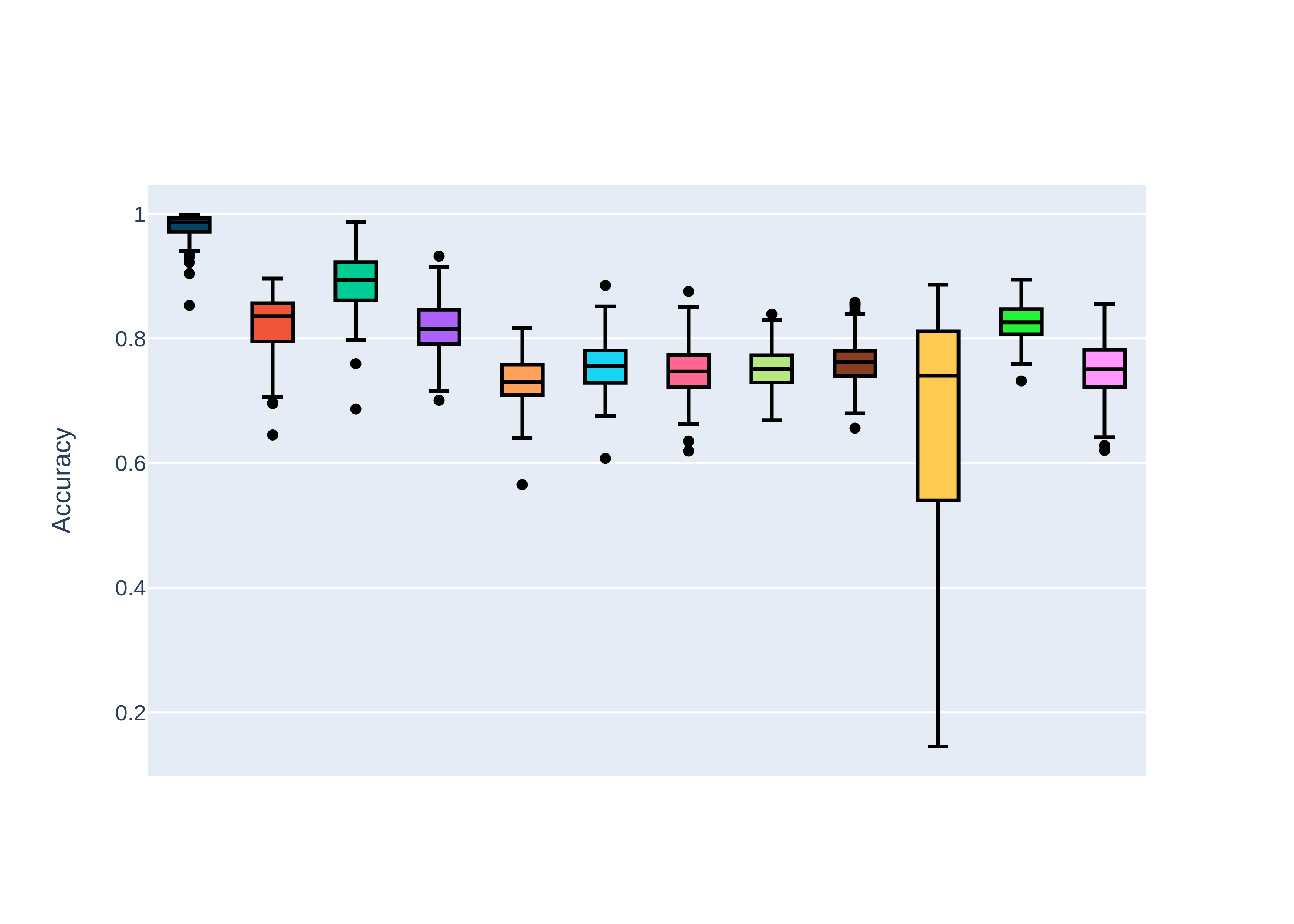}
            \captionsetup{justification=centering,font={footnotesize}}
            \caption{CG6}
        \end{subfigure}\\
        \begin{subfigure}[b]{0.32\textwidth}
            \includegraphics[trim = 60 400 400 500, clip, width=\textwidth]{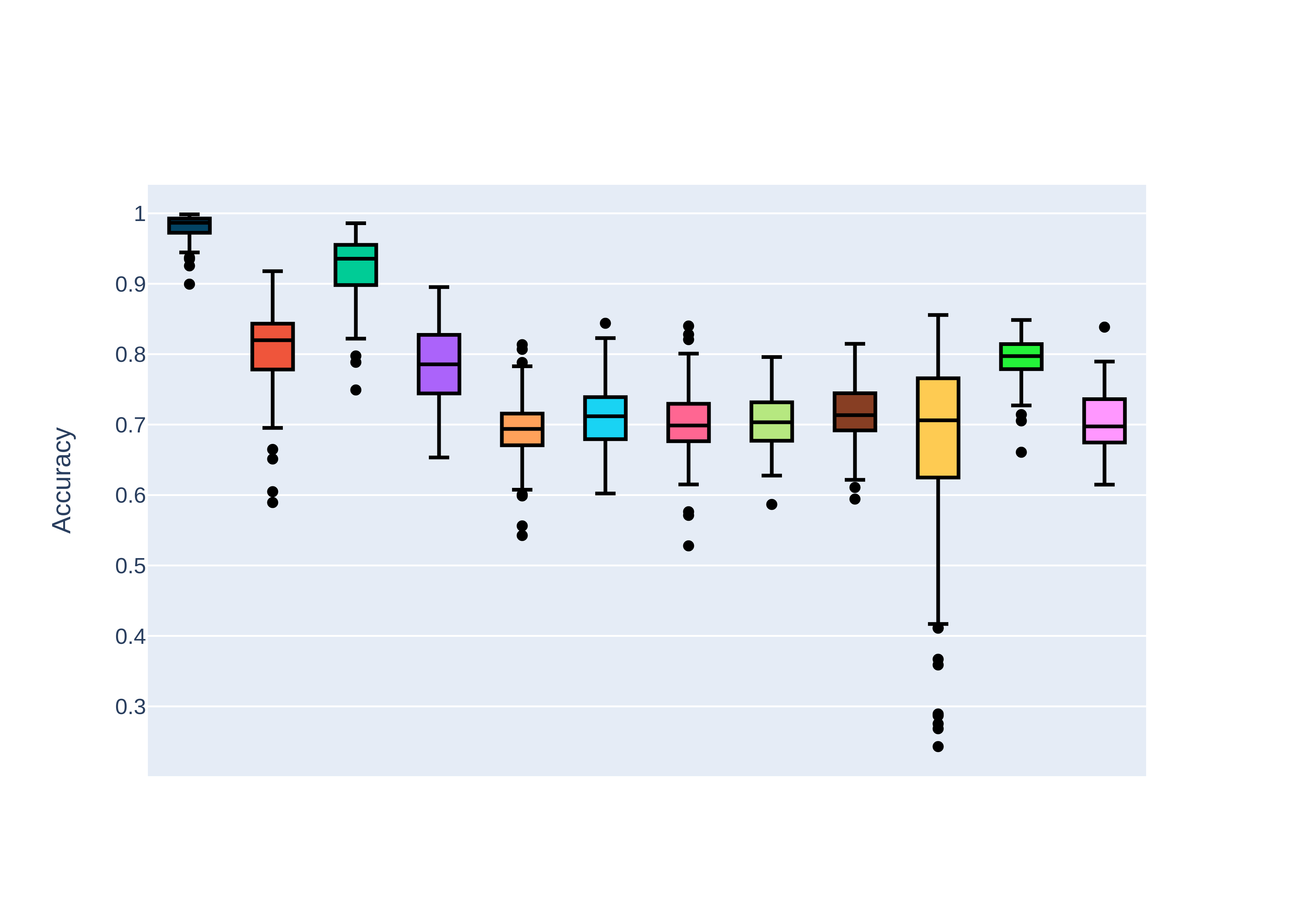}
            \captionsetup{justification=centering,font={footnotesize}}
            \caption{CG7}
        \end{subfigure}
        \begin{subfigure}[b]{0.32\textwidth}
            \includegraphics[trim = 60 400 400 500, clip, width=\textwidth]{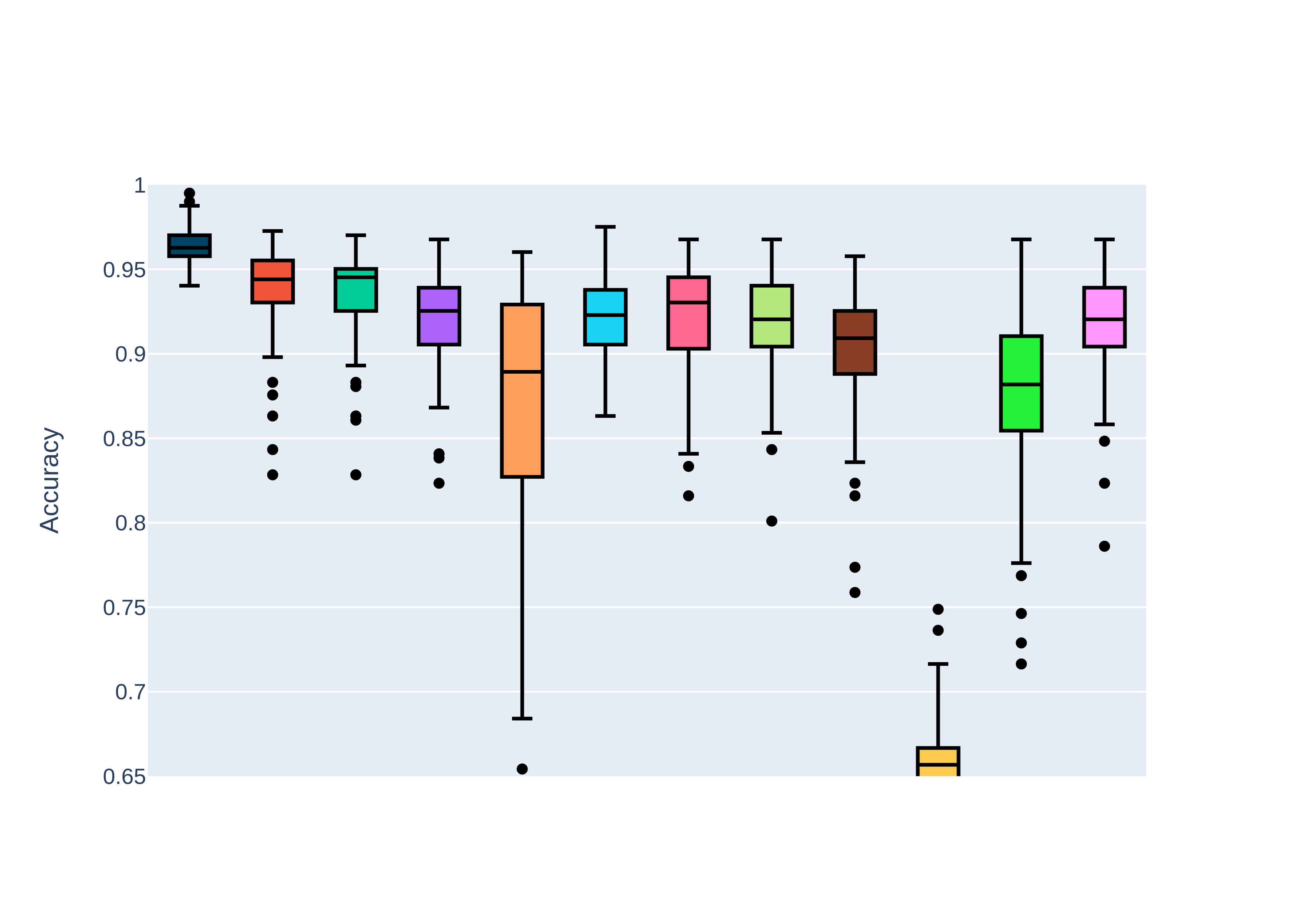}
            \captionsetup{justification=centering,font={footnotesize}}
            \caption{BCANCER}
        \end{subfigure}
        \begin{subfigure}[b]{0.32\textwidth}
            \includegraphics[trim = 60 400 400 500, clip, width=\textwidth]{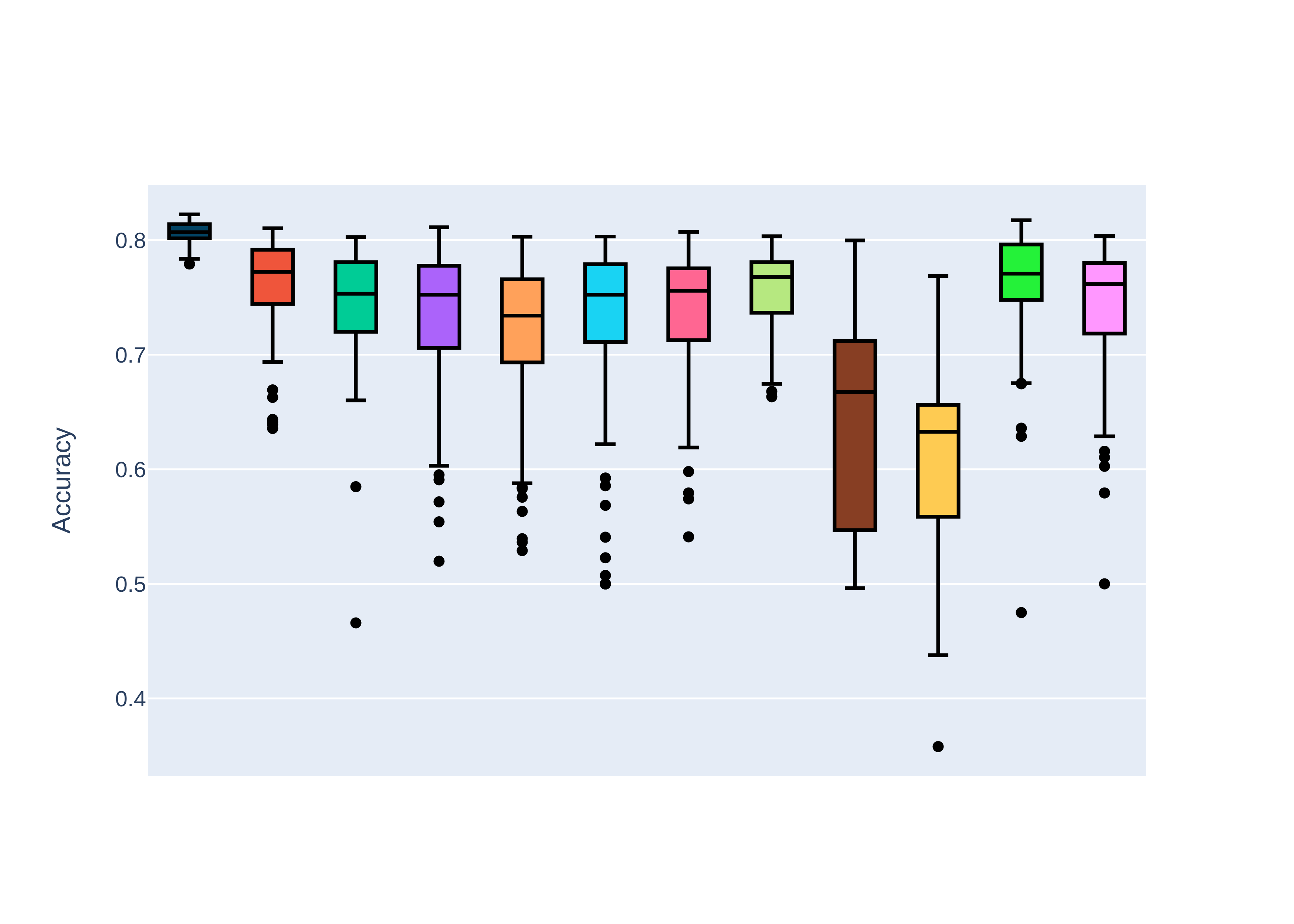}
            \captionsetup{justification=centering,font={footnotesize}}
            \caption{RAF}
        \end{subfigure}
    
    \end{adjustbox}
    \captionsetup{justification=centering,width=1.0\textwidth}
    \caption{Box plots: Prediction accuracy for synthetic ($|D_u|=10000)$) and real data over $D_U$. Colour and model correspondence is given in column `KEY' of Table \ref{tab:ulab_acc_synthetic}.}
    \label{fig:bplot_synthetic_performance}
    
\end{figure*}

\subsection{Real-world data}
\textbf{Datasets}. We use two real-world datasets: in the Breast Cancer Wisconsin dataset \cite{bcancer_wisconsin}, the goal is to classify tumour status as either malignant or benign. Using Tetrad software \cite{tetrad_software}, we search for the Markov Blanket using fast greedy equivalence search (Markov Blanket) (FGES-MB) \cite{fges_mb} with Degenerate Gaussian BIC score \cite{degenerate_dg_score}. From the Sachs cell-flow Cytometry dataset \cite{sachs_orig_paper}, we identify the Markov Blanket of binarised variable $RAF$. All datasets are originally given with no missing labels: hence, we create unlabelled partitions for SSL, using partition splits given in Table \ref{tab:real_partition_splits}. Partition allocation is randomised over $n=100$ trials, partition sizes are kept constant, and $P(Y=0)\approx 0.5$. 

\begin{figure}[H]
\vspace{-0.5cm}
\centering
\begin{subfigure}[t]{.48\linewidth}
\centering
\begin{tikzpicture}
	\begin{pgfonlayer}{nodelayer}
		\node [style=feature] (xc) at (-2, 0) {{$X_C$}};
		\node [style=feature] (y) at (0, 0) {{$Y$}};
		\node [style=feature] (xe) at (2, 0) {{$X_{E}$}};
		\node [style=feature] (xs) at (2, 1.5) {{$X_S$}};
	\end{pgfonlayer}
	\begin{pgfonlayer}{edgelayer}
	    \draw [-latex] (xc) to (y);
	    \draw [-latex] (y) to (xe);
		\draw [-latex] (xs) to (xe);
	\end{pgfonlayer}
\end{tikzpicture}
\captionsetup{justification=centering}
\caption{No confounders}
\end{subfigure}%
\begin{subfigure}[t]{.48\linewidth}
\centering
\begin{tikzpicture}
	\begin{pgfonlayer}{nodelayer}
		\node [style=feature] (xc) at (-2, 0) {{$X_C$}};
		\node [style=feature] (y) at (0, 0) {{$Y$}};
		\node [style=feature] (xe) at (2, 0) {{$X_{E}$}};
		\node [style=feature] (xs) at (2, 1.5) {{$X_S$}};
		\node [style=latent] (xi) at (0, 1.5) {{$X_I$}};
	\end{pgfonlayer}
	\begin{pgfonlayer}{edgelayer}
	    \draw [-latex] (xc) to (y);
	    \draw [-latex] (y) to (xe);
		\draw [-latex] (xs) to (xe);
		\draw [-latex] (xi) to (xc);		
		\draw [-latex] (xi) to (xs);
	\end{pgfonlayer}
\end{tikzpicture}
\captionsetup{justification=centering}
\caption{Latent confounder $X_I$}
\end{subfigure}
\captionsetup{justification=centering}
\caption{$X_I$ is a latent confounder of $X_C,X_S$}
\label{fig:real_data_confounder}
\end{figure}
\textbf{SCM approach for real-world data}. When creating the hierarchical SCM model for real-world data, we assume that the full DAG is not known as root nodes cannot be identified. The true data-generating causal model may contain a latent confounding feature $X_I$. This is depicted in Figure \ref{fig:real_data_confounder}. In this instance, we elect to model all parent features $X_C$ and spouse features $X_S$ as a single joint random variable. Hence, the generative model encompasses the possible correlation between any $X_C,X_S$.

\textbf{SSL performance on real-world data}. Both of our approaches, disjoint CGAN-SSL and joint GCGAN-SSL, outperform benchmark generative approaches on both datasets and improve on the baseline. GCGAN-SSL performs worse than the disjoint approach CGAN-SSL on real-world datasets, particularly for Sachs (RAF). This may be due to lack of realism in the dataset: we created a binary variable by introducing an arbitrary cutoff at the median value of RAF. Regions of discontinuity or low-density are crucial in inferring the effect of $Y$ if there are some clusters in $P(X_E|X_C)$. However, owing to our arbitrary choice of labelling function, it is possible that $X_E$ does not change very much when the label changes. In this case, the disjoint method may implicitly learn a more accurate model because we expect it to be less sensitive to changes in $Y$ for unlabelled data.

\section{Conclusions}
Our primary goal in this work is to provide a unified procedure for semi-supervised classification over causal graphical models. While our results confirm previous insights that the relationship $Y\rightarrow X_E$ is crucial for SSL, we show that a structural model $f_{X_E}$ which encapsulates explicit causal relations between $Y,X_E$ and extra features $X_C,X_S$ can improve the classifier by exploiting information in the paired observations in the unlabelled dataset. Our results illustrate the crucial importance of causality in SSL, which we hope may serve as a basis for more detailed analysis in future works.

\bibliography{bibtex/bib/references.bib}{}

\bibliographystyle{IEEEtran}

\end{document}